\RequirePackage{fix-cm}
\documentclass[twocolumn]{svjour3}          
\pdfoutput=1
\smartqed  
\usepackage{graphicx}
\usepackage{natbib}
\usepackage{multirow}
\usepackage{amsmath}
\usepackage{amssymb}
\usepackage{bbm}
\usepackage{bm}
\usepackage{booktabs}
\usepackage{array}
\usepackage{blindtext}
\usepackage{comment}

\usepackage{pifont}

\usepackage{xcolor}
\usepackage{color, colortbl}
\definecolor{citecolor}{HTML}{0071bc}
\definecolor{tabhighlight}{HTML}{e5e5e5}
\usepackage{hyperref}
\hypersetup {colorlinks,citecolor=citecolor}

\usepackage{tabulary,xspace,makecell}
\usepackage{dsfont,fixmath,mathtools,nicefrac}

\usepackage{subfigure}
\usepackage{overpic}
\usepackage{color, colortbl}
\definecolor{deemph}{gray}{0.6}

\definecolor{GrayBG}{gray}{0.95}
\definecolor{Highlight}{HTML}{39b54a}  
\definecolor{Green}{HTML}{39b54a}
\definecolor{Purple}{RGB}{167,1,233}

\definecolor{demphcolor}{RGB}{90,90,90}

\newcolumntype{x}[1]{>{\centering\arraybackslash}p{#1pt}}

\newcommand{\smallsec}[1]{\vspace{0.5ex}\noindent\textbf{#1}\quad}
\DeclareUnicodeCharacter{2212}{-}

\newcolumntype{d}[1]{D{.}{.}{#1}}

\newlength\savewidth\newcommand\shline{\noalign{\global\savewidth\arrayrulewidth
  \global\arrayrulewidth 1pt}\hline\noalign{\global\arrayrulewidth\savewidth}}
\newcommand{\tablestyle}[2]{\setlength{\tabcolsep}{#1}\renewcommand{\arraystretch}{#2}\centering\footnotesize}

\newcommand{\cmark}{\checkmark} 

\newcommand{\etal}{\textit{et al.}}

\newcommand{\I}{\mathbf{I}}

\newcommand{\C}{\mathbf{C}}

\newcommand{\Sx}[1]{\mathbf{S}_{#1}}

\newcommand\crule[3][black]{\textcolor{#1}{\rule{#2}{#3}}}
\def\clegend#1#2{\crule[#1]{6pt}{6pt} {\color{#1} #2}}

\def\up#1{(\textcolor[rgb]{0,0.75,0.25}{$\uparrow${#1}})}

\newcommand{\bfsec}[1]{\vspace{2mm}\noindent\textbf{#1}}

\definecolor{water}{RGB}{0, 0, 255}
\definecolor{non-vegetated-ground}{RGB}{128, 128, 128}
\definecolor{low-vegetation}{RGB}{0, 128, 0}
\definecolor{tree}{RGB}{0, 255, 0}
\definecolor{buildings}{RGB}{128, 0, 0}
\definecolor{playgrounds}{RGB}{255, 0, 0}

\definecolor{non_damage}{RGB}{0, 237, 253}
\definecolor{minor_damage}{RGB}{0, 255, 0}
\definecolor{major_damage}{RGB}{237, 142, 0}
\definecolor{destroyed}{RGB}{255, 0, 0}

\definecolor{ImpSurface}{RGB}{96, 96, 96}
\definecolor{Agriculture}{RGB}{204, 204, 0}
\definecolor{Forest}{RGB}{0, 204, 0}
\definecolor{Wetlands}{RGB}{0, 0, 153}
\definecolor{Soil}{RGB}{153, 76, 0}
\definecolor{Water}{RGB}{0, 128, 255}

\definecolor{convcolor}{HTML}{412F8A}
\definecolor{resnetcolor}{HTML}{8DA0CB}
\definecolor{vitcolor}{HTML}{fc8e62}

\definecolor{indomain}{HTML}{DAE8FC}
\definecolor{outdomain}{HTML}{FFE38E}

\newcommand{\bp}[1]{\noindent\textbf{#1}}

\newcommand{\convcolor}[1]{\textcolor{convcolor}{#1}}
\newcommand{\vitcolor}[1]{\textcolor{vitcolor}{#1}}
\newcommand{\vb}{\vitcolor{$\mathbf{\circ}$\,}}
\newcommand{\cb}{\convcolor{$\bullet$\,}}

\usepackage{soul}
\soulregister\cite7
\soulregister\citep7
\soulregister\ref7

\definecolor{regular}{rgb}{1,1,1}
\definecolor{revision}{rgb}{1,1,0}
\sethlcolor{revision}

\newcolumntype{a}{>{\columncolor{yellow}}l}
\newcolumntype{b}{>{\columncolor{yellow}}c}

\journalname{IJCV}
\begin{document}
\sloppy

\title{Single-Temporal Supervised Learning for Universal Remote Sensing Change Detection}


\author{Zhuo Zheng \and
        Yanfei Zhong \and
        Ailong Ma \and
        Liangpei Zhang
}


\institute{Zhuo Zheng \at
              LIESMARS, Wuhan University, China \\
              Department of Computer Science, Stanford University, USA \\
              \email{zhuozheng@cs.stanford.edu}
           \and
           Yanfei Zhong (corresponding author) \at
              LIESMARS, Wuhan University, China \\
              \email{zhongyanfei@whu.edu.cn}
           \and
           Ailong Ma \at
              LIESMARS, Wuhan University, China \\
              \email{maailong007@whu.edu.cn}
           \and
           Liangpei Zhang \at
              LIESMARS, Wuhan University, China \\
              \email{zlp62@whu.edu.cn}
}

\date{Received: date / Accepted: date}

\maketitle


\begin{abstract}
    Bitemporal supervised learning paradigm always dominates remote sensing change detection using numerous labeled bitemporal image pairs, especially for high spatial resolution (HSR) remote sensing imagery.
    However, it is very expensive and labor-intensive to label change regions in large-scale bitemporal HSR remote sensing image pairs.
    In this paper, we propose single-temporal supervised learning (STAR) for universal remote sensing change detection from a new perspective of exploiting changes between unpaired images as supervisory signals.
    STAR enables us to train a high-accuracy change detector only using unpaired labeled images and can generalize to real-world bitemporal image pairs.
    To demonstrate the flexibility and scalability of STAR, we design a simple yet unified change detector, termed ChangeStar2, capable of addressing binary change detection, object change detection, and semantic change detection in one architecture.
    ChangeStar2 achieves state-of-the-art performances on eight public remote sensing change detection datasets, covering above two supervised settings, multiple change types, multiple scenarios.
    The code is available at \url{https://github.com/Z-Zheng/pytorch-change-models}.
\end{abstract}

\section{Introduction} \label{sec:intro}

Change detection with multi-temporal high-resolution remote sensing imagery is a fundamental Earth vision task, which aims to provide prompt and accurate per-pixel change information on land surface for extensive real-world applications, such as urban expansion, sustainable development measurement, environmental monitoring, and disaster assessment~\citep{hussain2013change, daudt2018urban, mahdavi2019polsar, gupta2019creating, changeos}.

\begin{figure}[htb]
    \centering
    \subfigure[Conventional Bitemporal Supervised Learning]{
       \begin{minipage}[b]{\linewidth}
          \includegraphics[width=\linewidth]{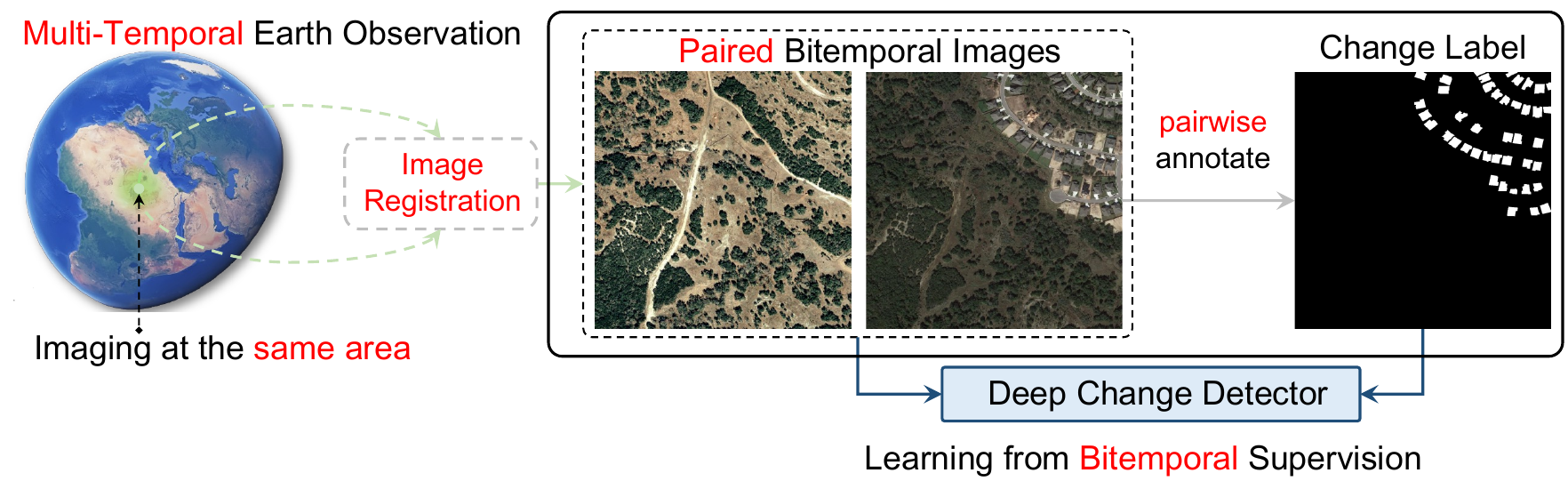}
       \end{minipage}
    }
    \subfigure[STAR: Single-Temporal supervised leARning]{
       \begin{minipage}[b]{\linewidth}
          \includegraphics[width=\linewidth]{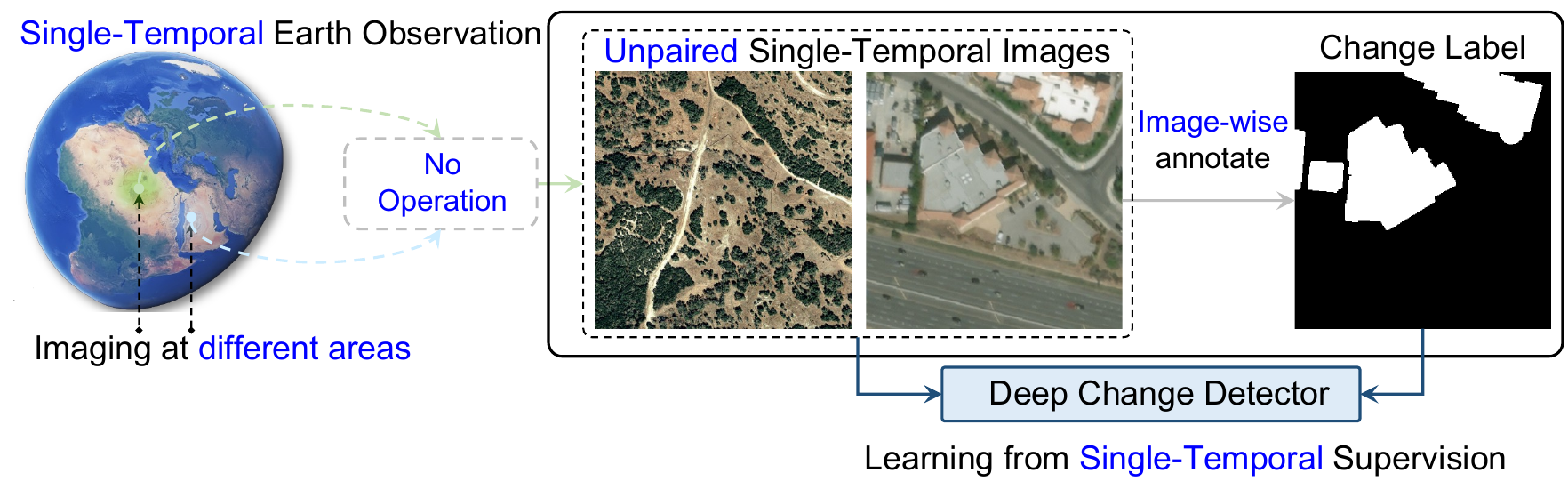}
       \end{minipage}
    }
    \caption{Comparison of conventional bitemporal supervised learning and the proposed single-temporal supervised learning for object change detection.
    By exploiting object changes in arbitrary image pairs as the supervisory signals, STAR makes it possible to learn a change detector from unpaired single-temporal images.
    }
    \label{fig:comp}
 \end{figure}

The state-of-the-art (SOTA) change detection methods are based on deep Siamese networks~\citep{bromley1993signature}, especially Siamese ConvNets~\citep{daudt2018fully, zheng2021change}.
Learning a deep Siamese network-based change detector needs a large number of labeled bitemporal image pairs for bitemporal supervision, as shown in Fig.~\ref{fig:comp} (a).
However, labeling large-scale and high-quality bitemporal high-resolution remote sensing image pairs is very expensive and time-consuming because of the extensive coverage of remote sensing images.
This problem significantly limits the real-world applications of the change detection technique.

 We observed that the importance of labeled bitemporal image pairs lies in that the change detector needs paired semantic information to define positive and negative samples of change detection.
 These positive and negative samples are usually determined by whether the pixels at two different times have different semantic labels in the same geographical area.
 The semantics of bitemporal pixels controls the label assignment, while the positional consistency condition\footnote{The bitemporal pixels should be at the same geographical position.} is only used to guarantee independent and identically distributed (i.i.d.) training and inference.
 It is conceivable that change is everywhere, especially between unpaired images, if we relax the positional consistency condition to define positive and negative samples.

 In this paper, we propose a single-temporal supervised learning to bypass the problem of collecting paired images by exploiting changes between unpaired images as supervisory signals, as shown in Fig.~\ref{fig:comp} (b).
 This approach enables us to train a high-accuracy change detector using unpaired images and generalize to real-world bitemporal image pairs at the inference stage.
 Because it only needs single-temporal semantic segmentation labels to generate change supervisory signals, we refer to our approach as \textit{Single-Temporal supervised leARning} (STAR).
 
 Conditioned by the same geographical area, bitemporal supervised learning can avoid many positive samples that are out-of-distribution in the entire bitemporal image space, whereas this is both an opportunity and a challenge for the STAR.
 These out-of-distribution samples make the change detector driven by STAR more potential to obtain better generalization.
 Meanwhile, they also cause the overfitting problem to make the model learn biased representation.
 To alleviate this problem, we explore an inductive bias: temporal symmetry and leverage it to constraint the representation learning for the change detector.
 
 To demonstrate the effectiveness of STAR, we design a simple yet unified change detector, \textit{ChangeStar2}, which follows the modular design and is made up of an arbitrary deep semantic segmentation model and ChangeMixin2 driven by STAR.
 ChangeMixin2 is designed to enable an arbitrary deep semantic segmentation model to detect object change.
 This allows ChangeStar2 to reuse excellent semantic segmentation architectures to assist in change detection without extra specific architecture design, which bridges the gap between semantic segmentation and change detection.
 
 The main contributions of this paper are summarized as follows:
 \begin{itemize}
    \item To fundamentally alleviate the problem of collecting paired labeled images, we proposed single-temporal supervised learning (STAR) to enable change detectors to learn from unpaired labeled images.
    \item To further stabilize the learning, we explore and leverage an inductive bias, temporal symmetry, to alleviate the overfitting problem caused by the absence of a positional consistency condition in unpaired images.
    \item To demonstrate the flexibility and scalability of STAR, we further proposed a simple yet unified change detector, termed ChangeStar2, which is capable of addressing binary change detection, object change detection, and semantic change detection.
    \item We demonstrated that ChangeStar2 performs SOTA on seven public (binary, object, semantic) change detection datasets, including the typical bitemporal supervised setting and in-domain/out-domain single-temporal supervised settings.
 \end{itemize}

This paper is an extended version of our previous work \citep{zheng2021change} accepted by ICCV 2021.
This work substantially extends the previous work in three aspects:

\noindent\textbf{1) Generalized to Semantic Change Detection.}
Beyond learning to detect object change via single-temporal supervision, generalized STAR enables change detectors to learn how to detect the semantic change by replacing the original \texttt{xor} label assigner with the semantic comparison label assigner.
This extension makes STAR become a unified change representation learning algorithm using unpaired images.

\noindent\textbf{2) Faster Convergence and Stronger Performance.}
According to our experimental analysis, we reveal that the reason of slow early convergence lies in the implicit temporal symmetry modeling of the temporal swap module.
To accelerate the convergence, we further propose the temporal difference network to improve ChangeMixin, which provides a learning-free surrogate of the temporal symmetric change representation to model temporal symmetry explicitly.
Also, it has been revealed that false positives caused by the absence of negative examples (unchanged objects) are the main error source in STAR.
To alleviate this problem, we propose the stochastic self-contrast strategy to add synthetic negative examples for a balanced learning procedure probabilistically.

\noindent\textbf{3) Unified Change Detection Architecture.}
To unleash the flexibility and scalability of STAR, we propose ChangeStar2 architecture towards universal remote sensing change detection.
ChangeStar2 is capable of addressing any change detection tasks (binary, object, semantic) and can learn from bitemporal supervision or single-temporal supervision.
We also scale ChangeStar2 with backbone networks (ConvNet, Transformer) of varying sizes to build a family of change detectors to meet multiple application scenarios, i.e., lightweight and fast change detectors for disaster response and heavy but accurate change detectors for large-scale Earth mapping.

\section{Related Work}

\bfsec{Object Change Detection.}
Different from general remote sensing change detection \citep{singh1989review}, object change detection is an object-centric change detection, which aims to answer the question whether the object of interest has been changed.
By the type of change, object change detection can be divided into two categories: binary object change detection, i.e., building change detection \citep{ji2018fully, chen2020spatial}, and semantic object change detection, i.e., building damage assessment \citep{gupta2019creating}, land cover change detection \citep{tian2020hiucd}.
Binary object change detection is a fundamental problem for object change detection.
The conference version \citep{zheng2021change} mainly focuses on binary object change detection.
In this work, we extend it to semantic (object) change detection toward generalized single-temporal supervised learning.

\bfsec{Semantic Change Detection.}
Apart from focusing on the occurrence of change, semantic change detection also focuses on the direction of change, i.e., ``from-to'' information, which also requires change detectors to obtain pre-change and post-change semantics.
Daudt~\etal \citep{daudt2019multitask} proposed a multi-task U-Net-like architecture to predict bitemporal semantic maps and a binary change map.
ChangeMask \citep{zheng2022changemask}, as a deep multi-task encoder-transformer-decoder architecture, was proposed to integrate with domain knowledge, i.e., temporal symmetry \citep{zheng2021change} and semantic-change causal relationship, to address inherent limitations of conventional multi-task architecture further.
To further consider semantic consistency and spatial context, Bi-SRNet \citep{ding2022bi} was proposed with self-attention and cross-attention layers to enhance convolution features with long-range context modeling.
Beyond detecting semantic changes on bitemporal image pair, DynamicEarthNet \citep{toker2022dynamicearthnet} poses a new challenging task to detect semantic changes on time-series satellite images, which needs time-continuous dense annotation to provide supervisory signals.
The annotations of time-series semantic change detection are much more expensive and time-consuming than object change detection.
Our proposed single-temporal supervised learning enables the model to learn to detect real-world bitemporal semantic changes from a time-agnostic single-image with its per-pixel semantic labels, significantly reducing the requirement of dense paired or time-continuous annotations.

\bfsec{Bitemporal Supervised Learning.}
So far, the supervised change detection methods are based on bitemporal supervised learning, which needs change labels from bitemporal remote sensing images of the same geographical area.
Although there are many datasets for change detection \citep{benedek2009change,bourdis2011constrained,fujita2017damage, lebedev2018change, ji2018fully, daudt2018urban, daudt2019multitask, chen2020spatial, tian2020hiucd}, their scales are still limited to meet the deep learning model.
Because paired or time-continuous annotation is very expensive and time-consuming.
Therefore, a more label-efficient learning algorithm for the change detector is necessary for real-world applications at scale.

\bfsec{Deep ConvNet Change Detector.}
Towards HSR remote sensing geospatial object change detection, the dominant change detectors are based on Siamese networks, such as fully convolutional Siamese network (FC-Siam) \citep{daudt2018fully}.
FC-Siam adopted a weight-shared encoder to extract temporal-wise deep features and then used a temporal feature difference decoder to detect object change from the perspective of encoder-decoder architecture.
The further improvements mainly focus on three perspectives of the encoder, i.e., using pre-trained deep network as encoder \citep{chen2020spatial, zhang2020deeply}, the decoder, i.e., RNN-based decoders \citep{mou2018learning, chen2019change}, spatial-temporal attention-based decoders \citep{chen2020spatial,zhang2020deeply}, transformer as decoder \citep{chen2021remote} and the training strategy, i.e., deep supervision for multiple outputs \citep{peng2019end,zhang2020deeply}.
There are apparent redundant network architecture designs because these network architectures are motivated by modern semantic segmentation models.
Therefore, it is significantly important for the next-generation change detector to reuse modern semantic segmentation architectures.

\bfsec{Object Segmentation.}
An intuitive but effective single-temporal supervised change detection method is the deep post-classification comparison (DPCC), which can serve as a strong baseline with the help of the modern deep segmentation model \citep{toker2022dynamicearthnet}.
However, this method only simply treats the change detection task as the temporal-wise semantic segmentation task and ignores the temporal information modeling.

\section{Single-Temporal Supervised Learning}

\begin{figure}
    \centering
    \includegraphics[width=\linewidth]{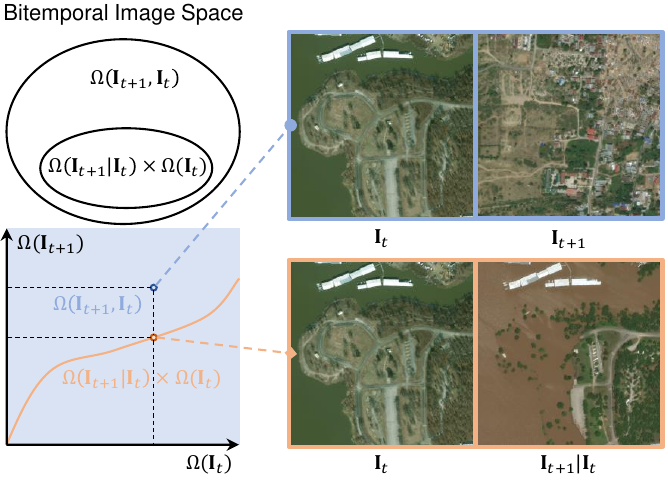}
    \caption{\textbf{Bitemporal Image Space}. Real-world bitemporal image space $\Omega(\I_{t+1}|\I_t)\times \Omega(\I_t)$ is a subset of the entire bitemporal image space $\Omega(\I_{t+1}, \I_t)$ because real-world bitemporal image pairs are conditioned by consistent spatial position, i.e., position consistency condition.
    }
    \label{fig:bis}
 \end{figure}

 \subsection{Rethinking Bitemporal Supervised Learning}
\begin{figure}[htb]
   \centering
   \subfigure[image $\I_t$]{
      \begin{minipage}[b]{0.3\linewidth}
         \includegraphics[width=\linewidth]{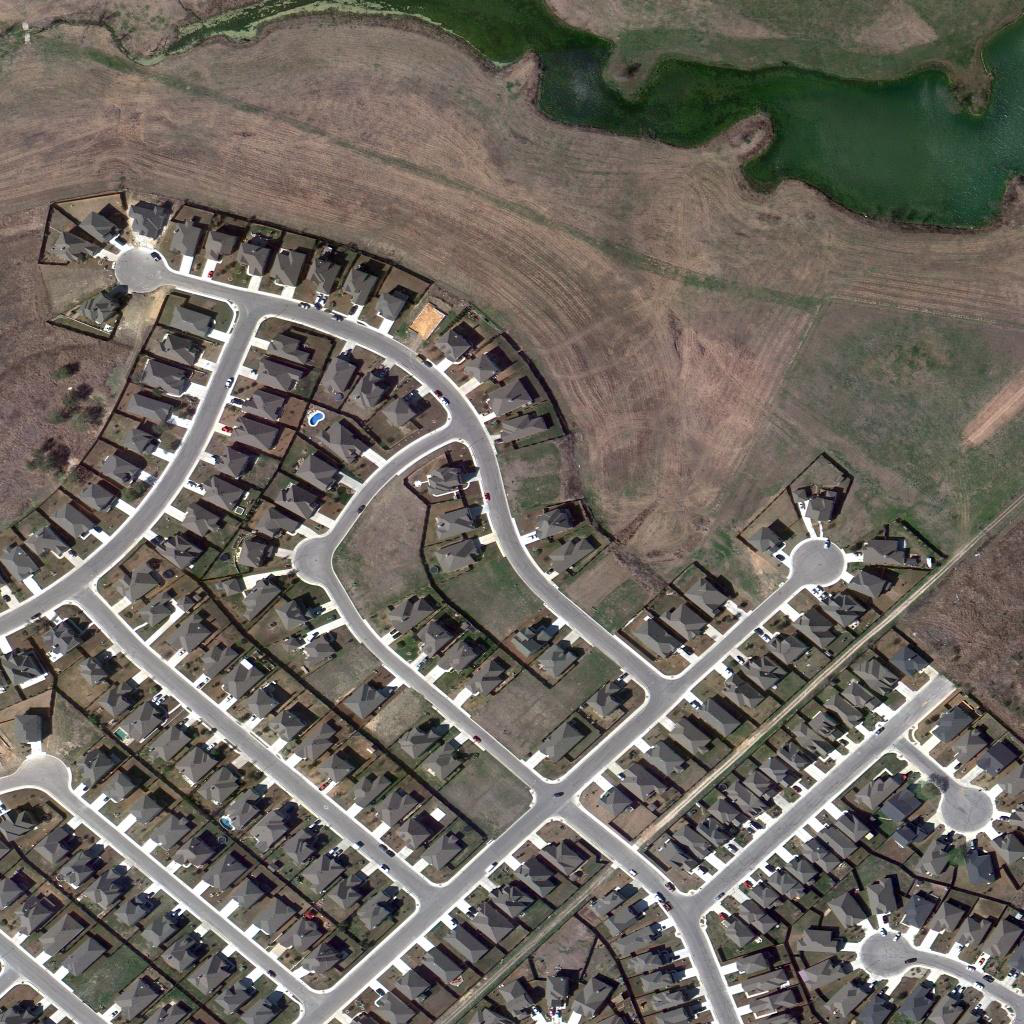}
      \end{minipage}
   }
   \subfigure[image $\I_{t+1}$]{
      \begin{minipage}[b]{0.3\linewidth}
         \includegraphics[width=\linewidth]{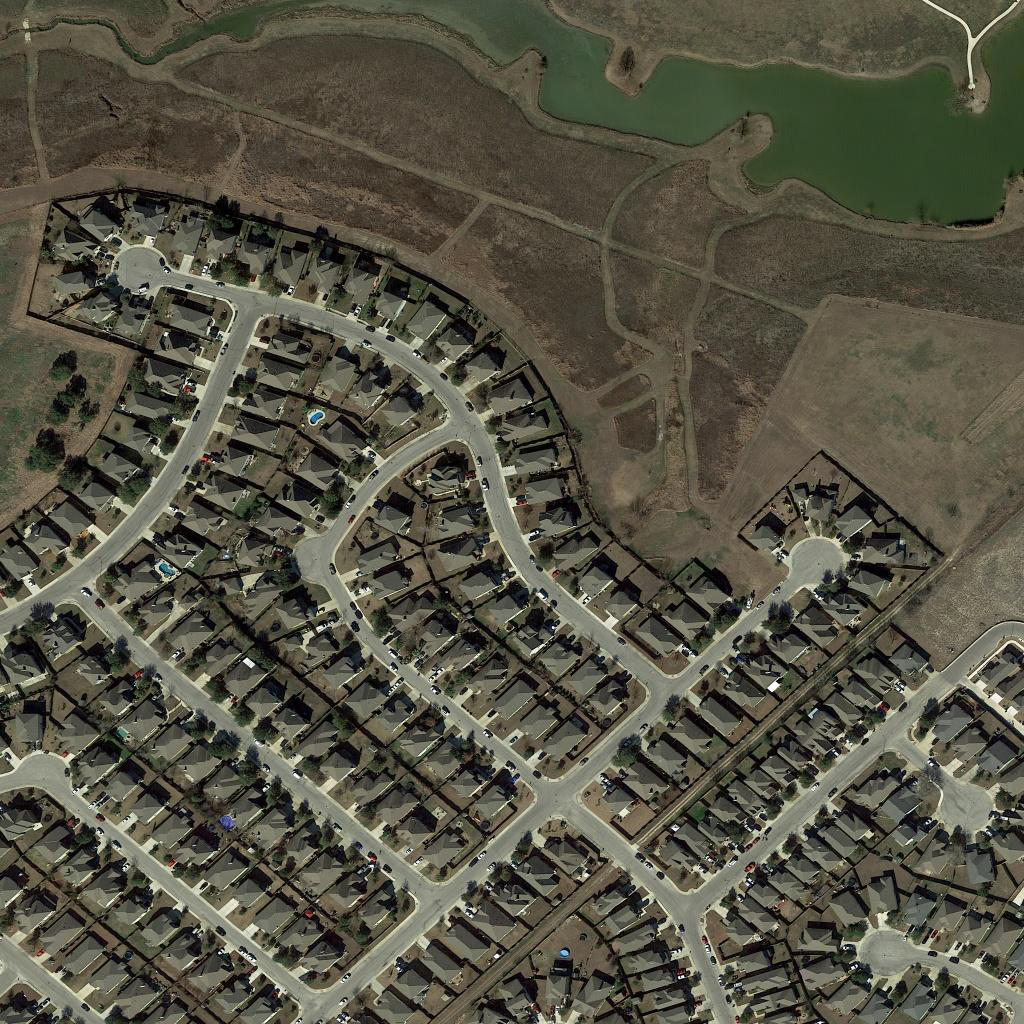}
      \end{minipage}
   }
   \subfigure[label $\C_{t\rightarrow t+1}$]{
      \begin{minipage}[b]{0.3\linewidth}
         \includegraphics[width=\linewidth]{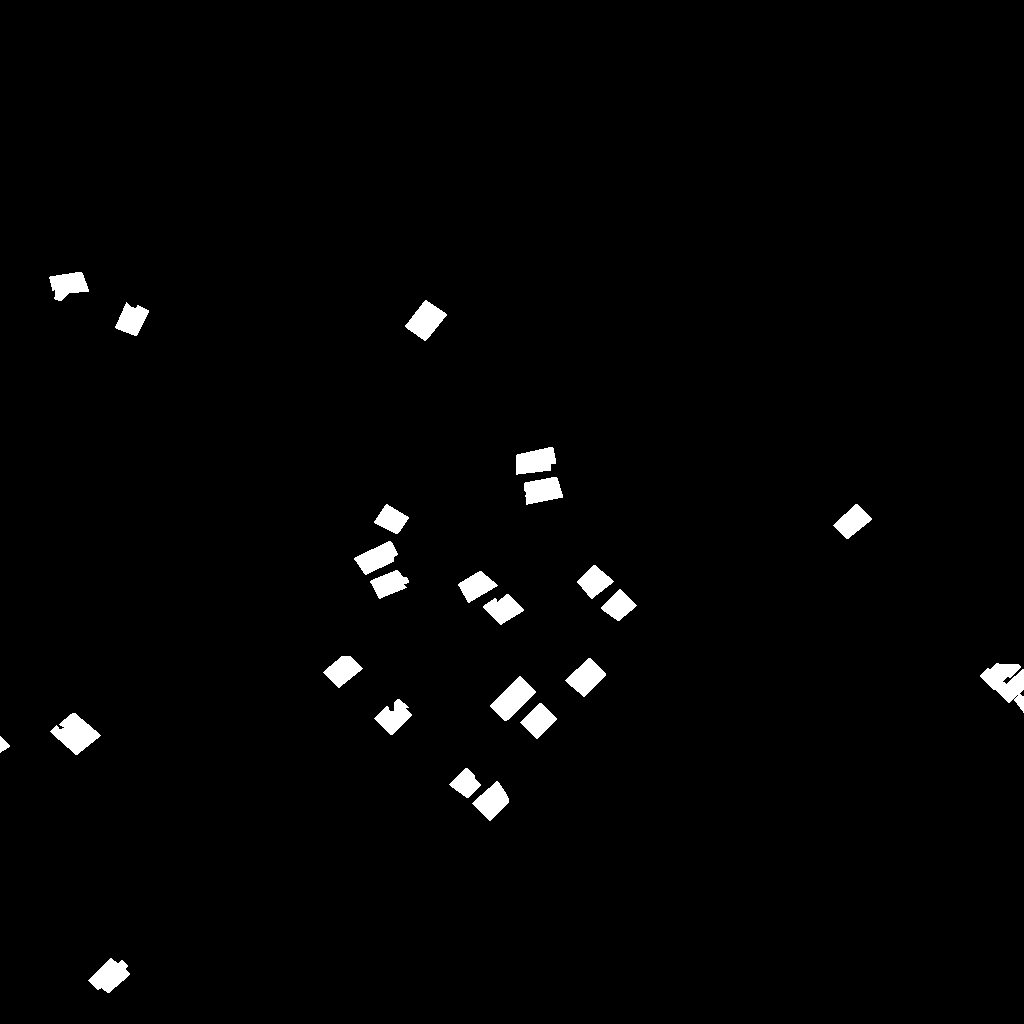}
      \end{minipage}
   }
   \caption{Training sample of bitemporal supervised object change detection. (a) the image at time $t$. (b) the image at time $t+1$. (c) change label representing the change happened the time period from $t$ to $t+1$.
      The image $\I_t$ should be spatially aligned with the image $\I_{t+1}$.
   }
   \label{fig:training_sample}
\end{figure}

 \bp{Original problem.}
 Learning a change detector with bitemporal supervision can be formulated as an optimization problem:
 \begin{equation}\label{eqn:org}
    \mathop{{\rm min}}\limits_{\theta} \mathbb{E}_{\I_t\sim p(\I_t),\I_{t+1}\sim p(\I_{t+1}|\I_t)}[\mathcal{L}(\mathbf{F}_{\theta}(\I_t,\!\I_{t+1}), \C_{t\rightarrow t+1})]
 \end{equation}
 where $p(\I_t)$ denotes the image data distribution at the time $t$, and $p(\I_{t+1}|\I_t)$ denotes the conditional data distribution at the time $t+1$.
 This conditional data distribution means the image $\I_{t+1}$ is conditioned by the image $\I_t$ in spatial position, i.e., positional consistency condition.
 As illustrated in Fig.~\ref{fig:bis}, for each image sampled from $p(\I_t)$, there is a corresponding post-change image sampled from $p(\I_{t+1}|\I_t)$, i.e., this image pair is sampled from the real-world bitemporal image space, i.e., $\Omega(\I_{t+1}|\I_t)\times \Omega(\I_t)$, which is a subset of the entire bitemporal image space $\Omega(\I_{t+1}, \I_t)$.
 The objective function $\mathcal{L}$ minimizes the cost between the prediction of the change detector $\mathbf{F}_\theta$ on bitemporal image pairs $\I_{t}, \I_{t+1} \in \mathbb{R}^{N\times C\times H\times W}$ and the change label $\mathbf{C}_{t\rightarrow t+1} \in \mathbb{R}^{N\times H\times W}$ representing the change occurring in the time period from $t$ to $t+1$.
 For example, Fig.~\ref{fig:training_sample} presents a training sample of bitemporal supervised object change detection.
 
 \bfsec{Relax positional consistency condition.}
 From Eq.~\ref{eqn:org}, we can find that the change label $\mathbf{C}_{t\rightarrow t+1}$ is the unique source of supervisory signals for change detection.
 To obtain $\mathbf{C}_{t\rightarrow t+1}$, paired semantic information is usually needed to define the positive and negative samples of change detection.
 However, paired semantic information is only related to the semantics of bitemporal pixels but is unrelated to their spatial positions.
 The same spatial position is only used to guarantee consistency between training and inference.
 If we relax the positional consistency condition, i.e., the image $\I_t$ is independent of the image $\I_{t+1}$ in terms of the spatial position, image pairs are directly sampled from the entire bitemporal image space $\Omega(\I_{t+1}, \I_t)$, as shown in Fig.~\ref{fig:bis}.
 The sampled image pair can be a composition of two arbitrary images.
 As such, the notation of time makes no sense here, thus, we adopt time-agnostic notation next for clearer formulation.
 Given two independent random remote sensing images $\I_i, \I_j\sim p(\I_i, \I_j)$, the original problem in Eq.~\ref{eqn:org} can be simplified as:
 \begin{equation}\label{eqn:relax}
    \mathop{{\rm min}}\limits_{\theta} \mathbb{E}_{\I_i, \I_j\sim p(\I_i, \I_j)}[\mathcal{L}(\mathbf{F}_{\theta}(\I_i, \I_j), \C_{i\rightarrow j})]
 \end{equation}
 where $\I_i, \I_j$ can be two unpaired images, and the change label $\C_{i\rightarrow j}:=\mathcal{A}(\Sx{i}, \Sx{j})$ can be more efficiently constructed from their semantic label $\Sx{i}, \Sx{j}$ by a label assigner $\mathcal{A}$.
 In this way, the hypothesis space of the model learned by Eq.~\ref{eqn:relax} is a superset of the hypothesis space of the model learned by Eq.~\ref{eqn:org}, which is allowed to detect object change in any context, including real-world bitemporal image pairs.

 \begin{figure*}[htb]
    \centering
    \begin{overpic}[width=\linewidth]{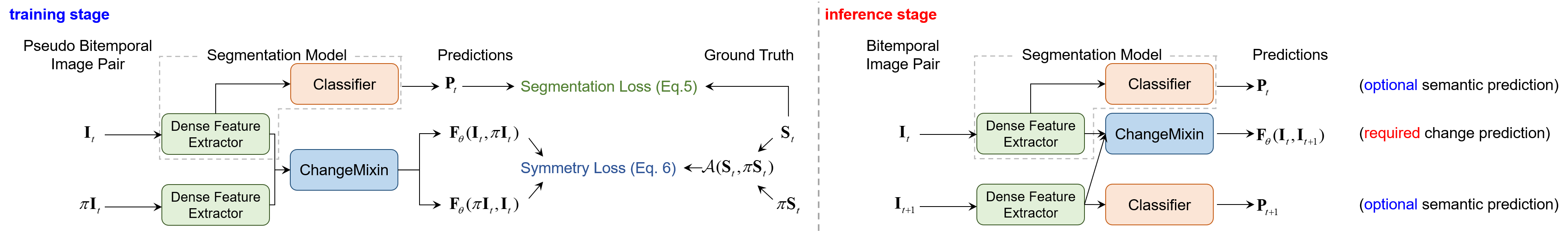}
    \end{overpic}
    \caption{\textbf{Overview of single-temporal supervised learning}.
       The network architecture of ChangeStar2 is made up of a dense feature extractor, a semantic classifier, and ChangeMixin.
       ChangeStar can be end-to-end trained by segmentation loss and symmetry loss with only single-temporal supervision.
       The modules with the same color share the weights.
    }
    \label{fig:overview}
 \end{figure*}
 
 \subsection{Generalized STAR}
 The key idea of single-temporal supervised learning, STAR, is to learn a generalizable change detector from arbitrary image pairs with only semantic labels via Eq.~\ref{eqn:relax}, as shown in Fig.~\ref{fig:overview}.
 To this end, we first introduce the pseudo bitemporal image pair, a key concept in STAR, to provide change supervisory signals.
 Based on the pseudo bitemporal image pair, bitemporal image pairs [$\I_t, \I_{t+1}$] in the original learning problem (Eq.~\ref{eqn:org}) can be replaced with single-temporal images $\I_t$, thus the learning problem can be reformulated as:
 \begin{equation}\label{eqn:gstar}
    \mathop{{\rm min}}\limits_{\theta} \, \mathbb{E}_{\I_t\sim p(\I_t)}[\mathcal{L}(\mathbf{F}_{\theta}(\I_t, \pi\I_t), \C_{t\rightarrow \pi(t)})]
 \end{equation}
 where the pseudo bitemporal image pairs [$\I_t, \pi\I_t$] with their change label $\C_{i\rightarrow \pi(t)}$ provide single-temporal supervision.
 The subscript $t$ is only used to represent that the data is single-temporal.

 \subsubsection{Pseudo Bitemporal Image Pair}
\begin{figure}[htb]
    \centering
    \subfigure[$\I_t$]{
       \begin{minipage}[b]{0.3\linewidth}
          \includegraphics[width=\linewidth]{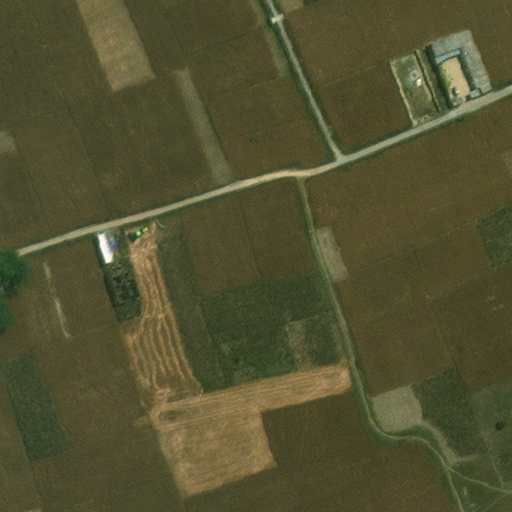}\vspace{4pt}
          \includegraphics[width=\linewidth]{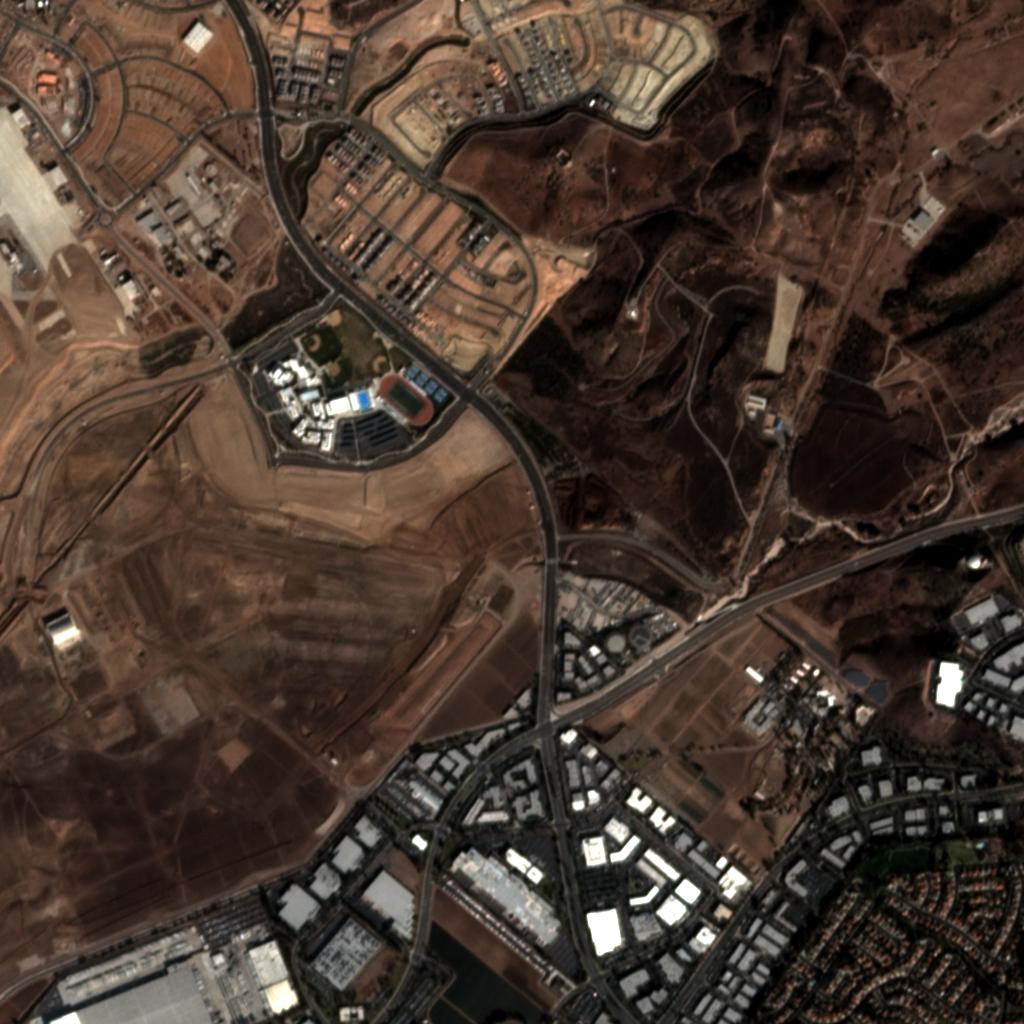}\vspace{4pt}
          \includegraphics[width=\linewidth]{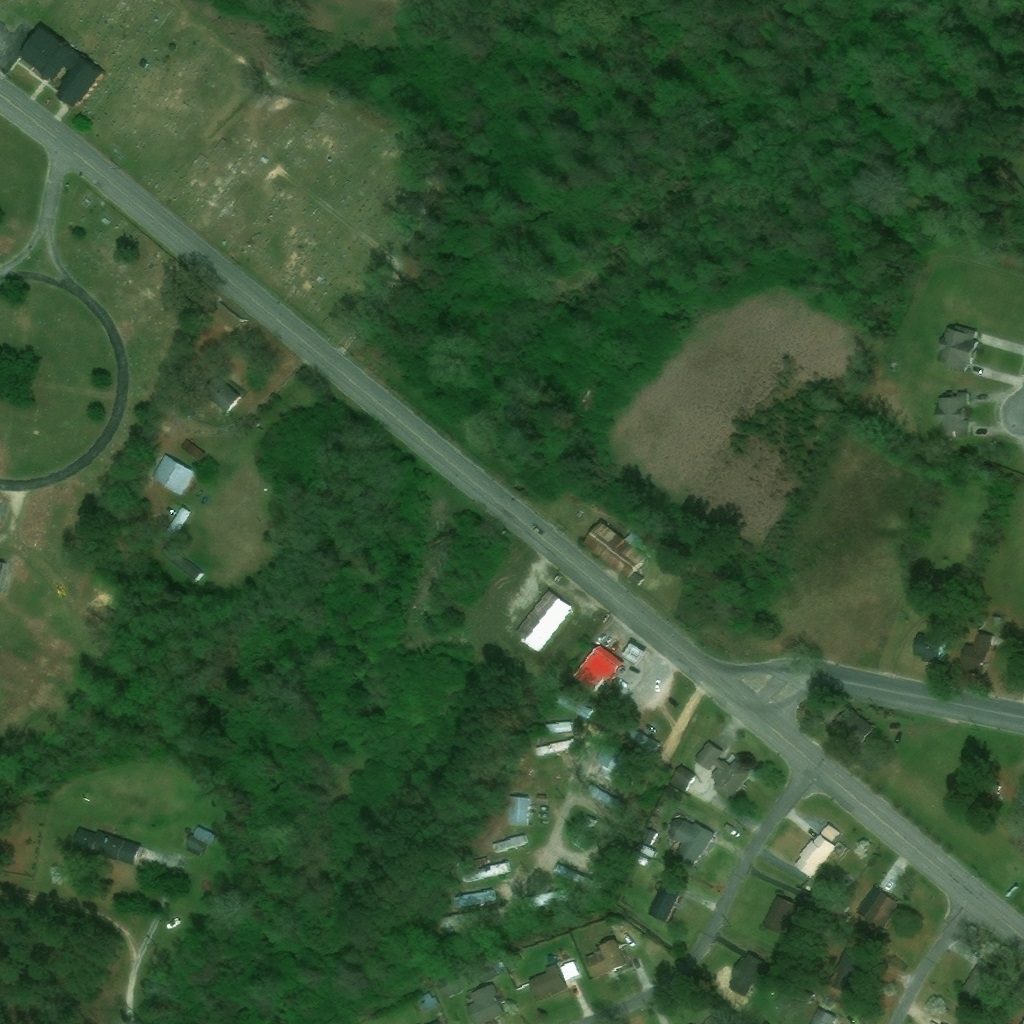}
 
       \end{minipage}
    }
    \subfigure[$\pi\I_t$]{
       \begin{minipage}[b]{0.3\linewidth}
          \includegraphics[width=\linewidth]{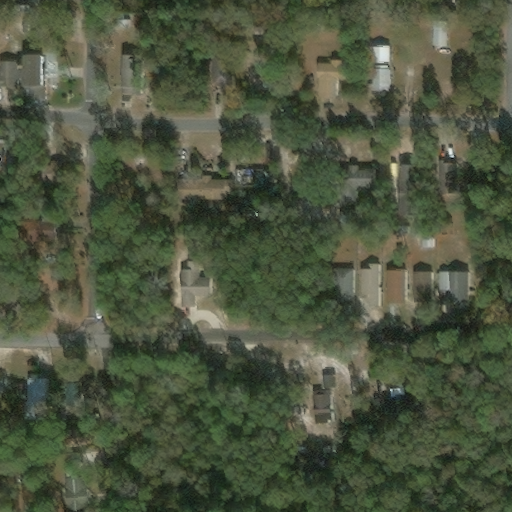}\vspace{4pt}
          \includegraphics[width=\linewidth]{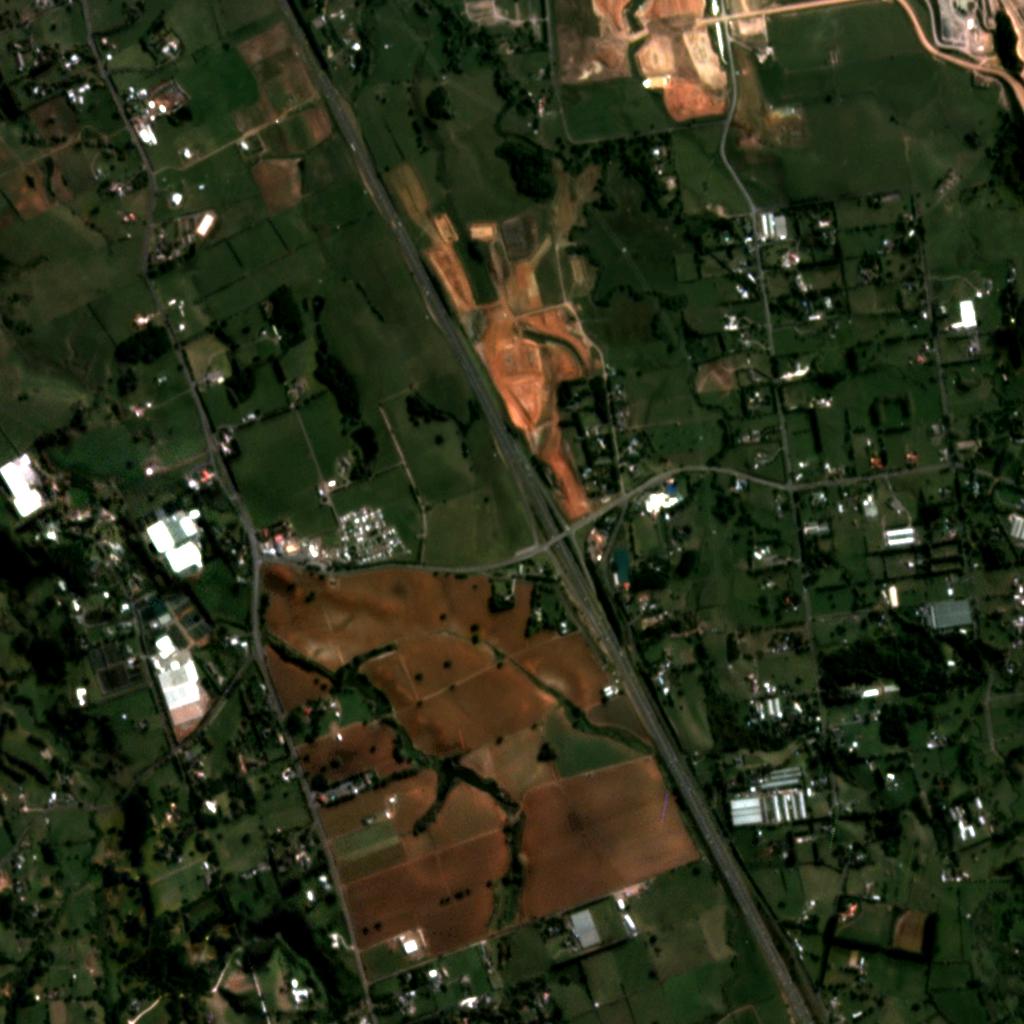}\vspace{4pt}
          \includegraphics[width=\linewidth]{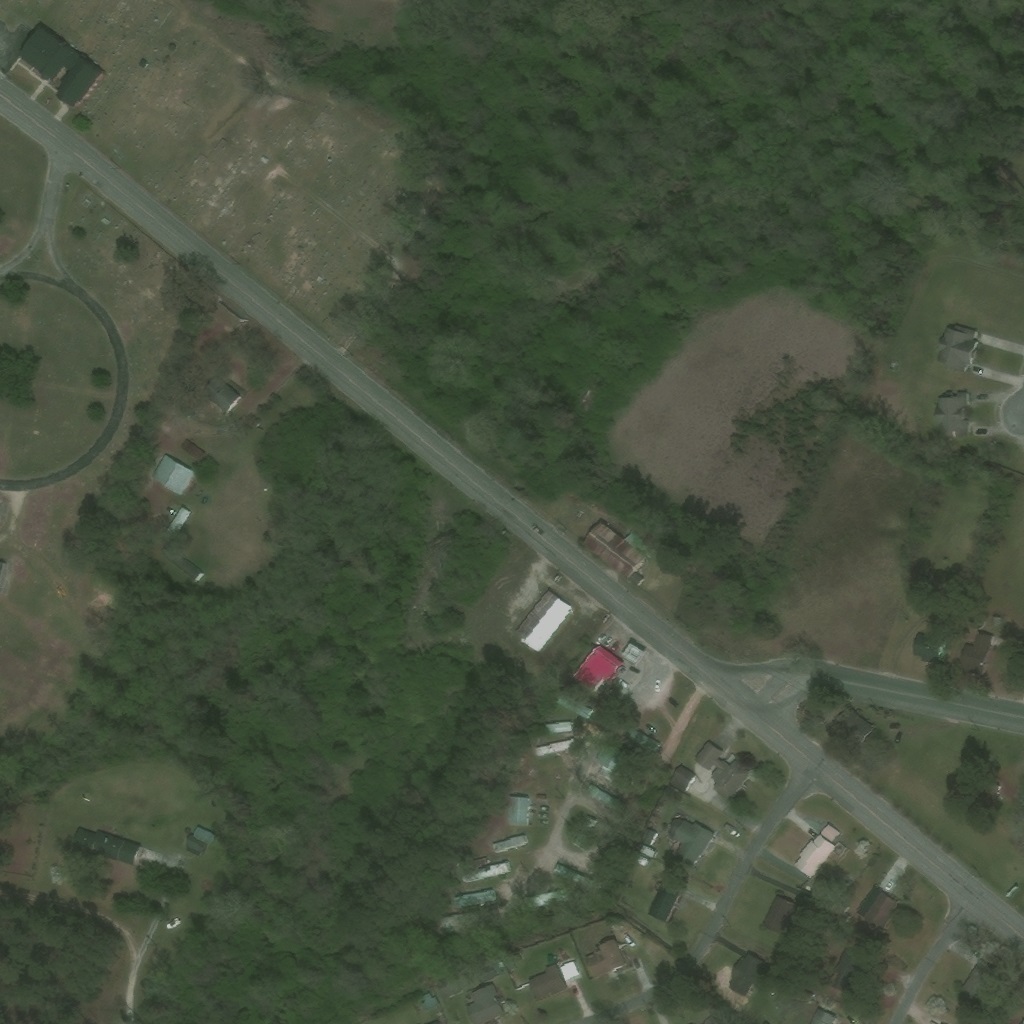}
 
       \end{minipage}
    }
    \subfigure[change label]{
       \begin{minipage}[b]{0.3\linewidth}
          \includegraphics[width=\linewidth]{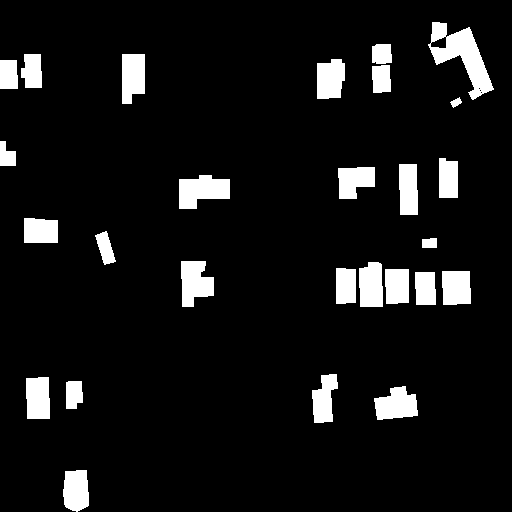}\vspace{4pt}
          \includegraphics[width=\linewidth]{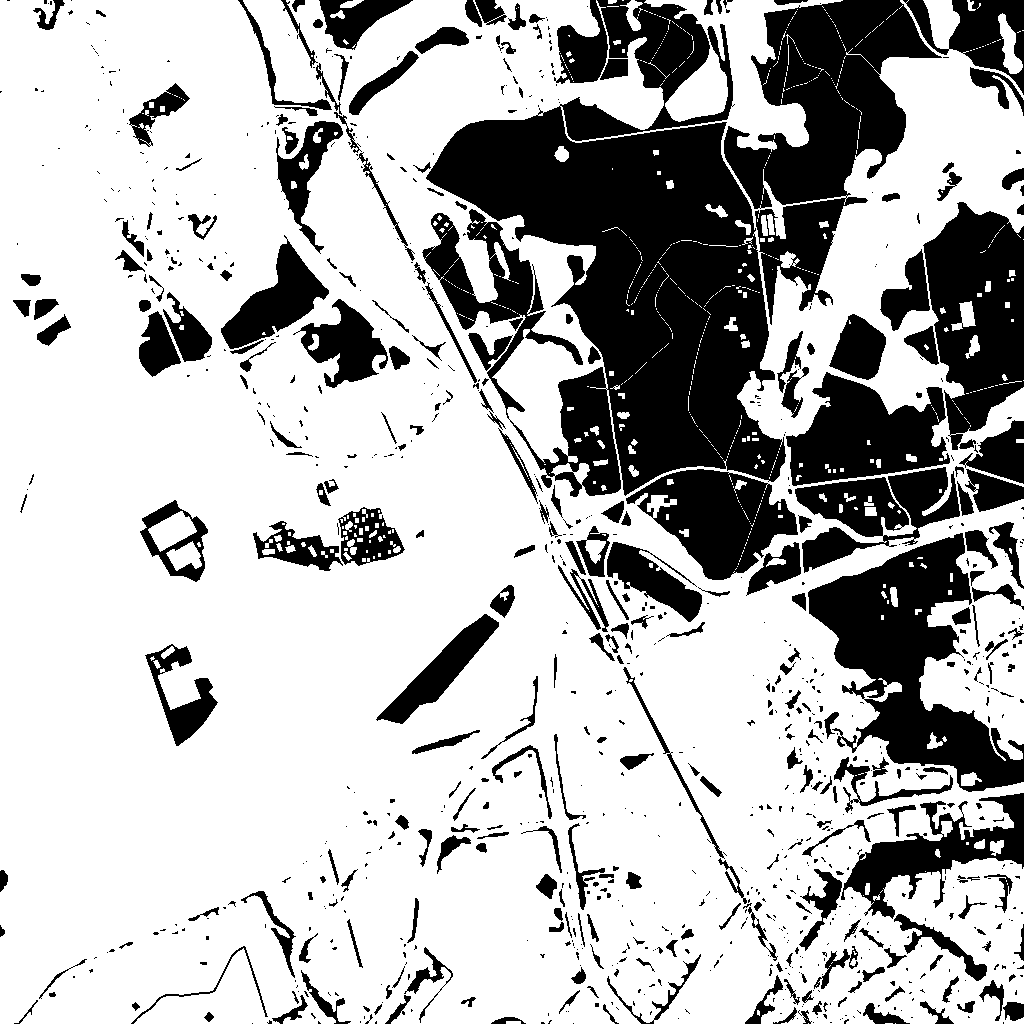}\vspace{4pt}
          \includegraphics[width=\linewidth]{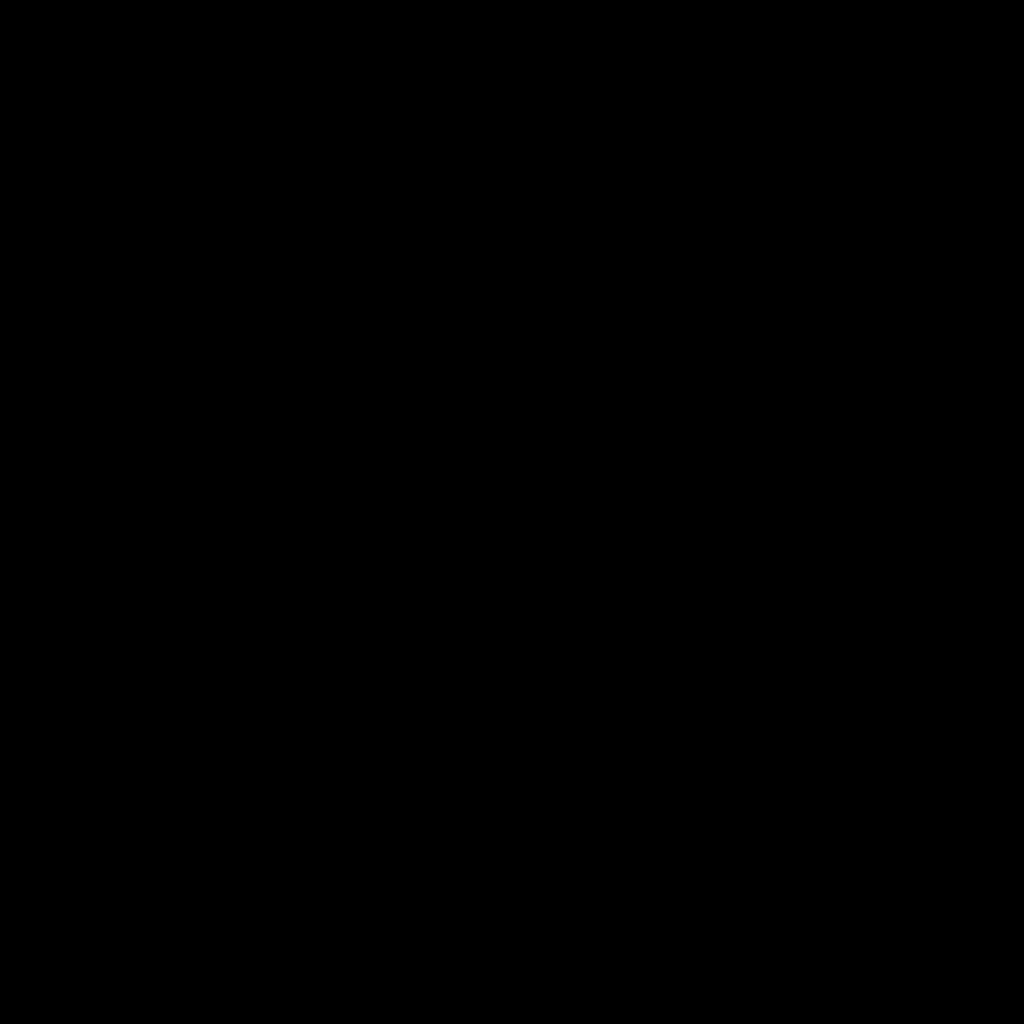}
 
       \end{minipage}
    }
    \caption{\textbf{Pseudo Bitemporal Image Pair}. There are three typical cases of the pseudo bitemporal image pair, i.e., object change detection (1st row), semantic change detection (2nd row), and generic self-contrast (3rd row).
       $\I_t$, $\pi\I_t$ are the original image sequence and the new image sequence generated by a random permutation~$\pi$.
       For change label, change regions are white and unchanged regions are black.
    }
    \label{fig:pseudo_bitemp}
 \end{figure}

 To provide change supervisory signals with single-temporal data, we first construct pseudo-bitmporal image pairs in a mini-batch and then assign labels to them during training.
 Fig.~\ref{fig:pseudo_bitemp} (c) demonstrates three typical cases of the pseudo-bitmporal image pair.

 \bfsec{Random Permutation in Mini-batch.}
 Given a mini-batch single-temporal images $\I_t$ with its semantic labels $\Sx{t}$, $\I_t$ can be seen as a sequence $\{\I^{(1)}_t, ...,  \I^{(n)}_t\}$.
 We use a random permutation $\pi \in S_n$ of this sequence to generate a new sequence $\pi\I_t$ to replace real-world next-time images $\I_{t+1}$, where $S_n$ denotes the all permutations of indices $\{1, ..., n\}$, and $\pi\I_t$ denotes the sequence $\{\I^{\pi(1)}_t, ...,  \I^{\pi(n)}_t\}$.
 Fig.~\ref{fig:pseudo_bitemp} (a) and (b) present the original sequence of three images and the new sequence in case of a mini-batch of three images.
 
 \bfsec{Label Assignment.}
 Different from manually paired dense labeling for bitemporal supervised learning, change labels are automatically generated by single-temporal semantic labels for STAR.
 The positive labels of object change are assigned to the pixel positions in which the object of interest only once appeared.
 If there are two object instances overlapped at pseudo bitemporal images, the pixel positions in the overlapping area are assigned as negative examples.
 The rest of the pixel positions are also assigned as negative examples.
 In this way, change labels $\C_{t \rightarrow t+1}$ in Eq.~\ref{eqn:org} can be replaced with $\mathcal{A}(\Sx{t}, \pi\Sx{t})=\mathbb{I}(\Sx{t}\neq \pi\Sx{t})$, where $\mathbb{I}$ denotes a element-wise indicator function. 
 This is a generalized label assigner that can be used for semantic change detection, compared to the original \texttt{xor}.

 \bfsec{Stochastic Self-Contrast.}
 The above label assignment always causes many false positives due to the absence of real-world negative examples \citep{zheng2021change}.
 To alleviate false positives in previous STAR, we further propose a stochastic self-contrast strategy to balance the learning procedure by introducing homologous negative examples from the same single-temporal image.
 For a pseudo bitemporal image pair [$\I^{(i)}_t, \I^{\pi(i)}_t$] in a mini-batch, there is a self-contrast probability~$p$ of replacing pseudo next-time image $\I^{\pi(i)}_t$ with an approximate next-time image $\widetilde{\I}_{t+1}$.
 This approximation is implemented by random color jitter, i.e., $\widetilde{\I}_{t+1}^{(i)}:={\rm aug}(\I^{(i)}_t)$, to guarantee spatially aligned negative examples.
 This self-contrast pseudo bitemporal image pair [$\I^{(i)}_t, \widetilde{\I}_{t+1}^{(i)}$] can provide sufficient synthetic negative examples, which is not only helpful to balance the learning procedure, but also provides consistency regularization \citep{bachman2014learning, sohn2020fixmatch} to alleviate the overfitting in the entire bitemporal image space.
 From the perspective of sampling, stochastic self-contrast provides a mechanism to control the sampling process of pseudo-bitemporal image pairs.
 Naive sampling via random permutation can be seen as sampling from the entire bitemporal image space $\Omega(\I_{t+1}, \I_t)$ to provide more generalized change supervisory signals.
 Self-contrast can be seen as sampling from approximate conditional space $\Omega(\widetilde{\I}_{t+1}|\I_t)\times \Omega(\I_t)$ to align data distribution towards real-world bitemporal image space.
 Stochastic self-contrast is a mixture of above two sample methods for a trade-off between more change supervisory signals and more real data distribution.

 \subsubsection{Multi-task Supervision}
 The overall objective function $\mathcal{L}$ is a multi-task objective function, which is used to sufficiently exploit single-temporal semantic labels for joint semantic segmentation and change detection, as follows:
 \begin{equation}
    \mathcal{L} = \mathcal{L}_{\texttt{seg}} + \mathcal{L}_{\texttt{change}}
 \end{equation}
 Generalized STAR adopts a decoupled semantic change representation \citep{zheng2022changemask}, which decouples the semantic change into the binary change and bitemporal semantics.
 In this way, generalized STAR can handle semantic change detection in the same way as binary change detection.

 \bfsec{Semantic Supervision.}
 For object segmentation, we adopt binary cross-entropy loss and dice loss \citep{milletari2016v} for the binary case and cross-entropy loss for the multi-class case to provide semantic supervision, respectively.

 \bfsec{Change Supervision by Temporal Symmetry.}
 Temporal symmetry is a mathematical property of binary change, which indicates that binary change is undirected, i.e. $\C_{t \rightarrow t+1} = \C_{t+1 \rightarrow t}$.
 Intuitively, the outputs of binary change detector on the bitemporal image pair should follow this property.
 This means that the binary change detector should not fit the temporal direction under the constraint of temporal symmetry.
 Motivated by this, we further propose symmetry loss for binary change detection, which is formulated as follows:
 \begin{equation}
    \begin{split}
       \mathcal{L}_{\rm change} =  \frac{1}{2}[&\mathcal{L}_{\rm binary}(\mathbf{F}_\theta(\I_t, \pi\I_t), \mathcal{A}(\Sx{t}, \pi\Sx{t})) \\
       + &\mathcal{L}_{\rm binary}(\mathbf{F}_\theta(\pi\I_t, \I_t), \mathcal{A}(\pi\Sx{t}, \Sx{t}))] \\
    \end{split}
 \end{equation}
 where $\mathcal{L}_{\rm binary}$ adopts the binary cross-entropy as default.
 For a special case of extremely imbalanced positive and negative samples, we adopt a compound loss of binary cross-entropy loss and soft dice loss.
 This symmetry loss features an inductive bias provided by temporal symmetry, which servers as a regularization term to alleviate the overfitting the temporal order in binary change detection.

 \section{Unified Change Detector: ChangeStar2}
 ChangeStar2 is a simple yet unified change detector capable of addressing binary change detection, object change detection, and semantic change detection, which is composed of a Siamese dense feature extractor and ChangeMixin2.
 This design is inspired by reusing the modern semantic segmentation architecture because semantic segmentation and change detection are both dense prediction tasks.
 To this end, we design ChangeMixin2 to enable any off-the-shelf deep semantic segmentation network to detect changes.
 Fig.~\ref{fig:arch} presents the overall architecture of ChangeStar2.
 Compared to ChangeMixin, ChangeMixin2 addresses the slow convergence problem while retaining architectural generality via a temporal difference network.

 \subsection{Siamese Dense Feature Extractor}
\begin{figure}
    \centering
    \includegraphics[width=.95\linewidth]{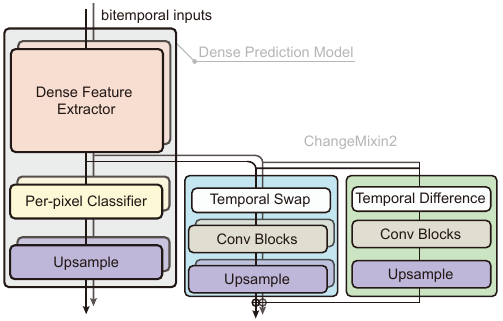}
    \caption{\textbf{ChangeStar2}.
    The network architecture of ChangeStar2 is made up of a dense prediction model and a ChangeMixin2 module.
    ChangeMixin2 has two branches, i.e., temporal swap network and temporal difference network.
    The temporal swap network contains a temporal swap module followed by several conv blocks.
    The temporal difference network contains a temporal difference module followed by several conv blocks.
    }
    \label{fig:arch}
 \end{figure}

 Siamese dense feature extractor can be simply implemented by a weight-sharing deep segmentation network without any architecture modification.
 Given a bitemporal image pair $\I_t, \I_{t+1} \in \mathbb{R}^{C_{\rm in}\times H\times W}$ as input, a off-the-shelf deep segmentation network, as the dense feature extractor, outputs a bitemporal high-resolution feature map pair $\mathbf{X}_{t}, \mathbf{X}_{t+1} \in \mathbb{R}^{C_{\rm out}\times \left\lfloor H/s\right\rfloor\times \left\lfloor W/s\right\rfloor}$, where $s$ is the output stride, e.g., $s=4$ in FarSeg \citep{zheng2020foreground} and $s=8$ in DeepLab v3 \citep{chen2017rethinking}.
 The feature maps $\mathbf{X}_{t}, \mathbf{X}_{t+1}$ are temporally-independent because the dense feature extractor is applied on each image separately via weight-sharing.
 The top block of a segmentation model is always a conv layer with $C$ filters as a per-pixel classifier, followed by an upsampling layer, where $C$ is the number of classes and the upsampling scale is equal to the output stride of the specific segmentation model.
 The feature map for change detection is only computed by the ``Dense Feature Extractor'' part of the entire model, as shown in Fig.~\ref{fig:arch} left.
 
 \subsection{ChangeMixin2}
 Beyond the original ChangeMixin, we further design a temporal difference network to address the remaining slow early convergence problem \citep{zheng2021change} via introducing explicit temporal symmetry.
 ChangeMixin2 can be seen as a composition of the temporal swap network and temporal difference network.
 We introduce these two networks next.
 
 \bfsec{Temporal Swap Network (implicit temporal symmetry)} is composed of a temporal swap module and a small FCN composed of $N$ 3$\times$3 conv layers, each with $d_c$ filters and each followed by BN and ReLU.
 Besides, a bilinear upsampling layer followed by a sigmoid activation is attached to output the binary predictions per pixel.
 The temporal swap (Eq.~\ref{eqn:tsm}) is responsible for temporal symmetry, providing an inductive bias in the network architecture, which takes bitemporal feature maps $\mathbf{X}_{t}, \mathbf{X}_{t+1}$ as input and then concatenates them along the channel axis in two different temporal permutations.
 \begin{equation}\label{eqn:tsm}
    \texttt{TSM}(\mathbf{X}_{t}, \mathbf{X}_{t+1}) = \texttt{cat}(\mathbf{X}_{t}, \mathbf{X}_{t+1}), \texttt{cat}(\mathbf{X}_{t+1}, \mathbf{X}_{t})
 \end{equation}
 During training, the small FCN is attached to each output of TSM and the weight of the small FCN is shared.
 During inference, the small FCN is only attached to the first output of TSM because we find that two outputs are approximately temporal-symmetric in the converged model.
 We use $N = 4$ and $d_c = 16$ for a better trade-off between speed and accuracy, following \citep{zheng2021change}.
 It can be seen that TSN implicitly models the temporal symmetry and relies on the convolution layers to learn change representation, which causes slow early convergence problems observed by our experiments (Fig.~\ref{fig:multitask_output}).
 To overcome this problem, we further propose the temporal difference network.
 
 \bfsec{Temporal Difference Network (explicit temporal symmetry)} is composed of a temporal difference module (TDM) and a small FCN projector.
 TDM is a parameter-free symmetric aggregation function (e.g. absolute difference, Hadamard product), which aims to provide a learning-free surrogate for temporal symmetric change representation, thus explicitly modeling temporal symmetry.
 We adopt the absolute difference to implement TDM by default for empirical faster convergence.
 With the temporal difference network, the temporal swap network only needs to progressively learn the residual change representation rather than directly approximating it.
 To this end, we further append a small FCN projector on TDM to align the channel dimension of the feature maps for residual feature fusion.
 Thus, the output feature map $\mathbf{X}_{t\rightarrow t+1}$ of ChangeMixin2 can be computed as:
 \begin{equation}\label{eqn:adbn}
    \mathbf{X}_{t\rightarrow t+1} = \mathbf{X}^{\rm tdn}_{t\leftrightarrow t+1} + \mathbf{X}^{\rm tsn}_{t\rightarrow t+1}
 \end{equation}
 where $\mathbf{X}^{\rm tdn}_{t\leftrightarrow t+1}$ denotes the temporal symmetric feature map from the temporal difference network, and $\mathbf{X}^{\rm tsn}_{t\rightarrow t+1}$ denotes the feature map from the temporal swap network in the temporal order of $t\rightarrow t+1$.

\subsection{Concept Comparison to Existing Change Detectors}
Regarding network architecture, ChangeStar and ChangeStar2 feature the concept of ``reuse modern semantic segmentation architecture''.
This is also one of the main differences compared to existing change detectors.
We can reuse this architecture to better model spatial context and semantics.
Also, thanks to the idea of architecture reuse, they can be used for various change detection tasks without any architectural modification.
The second difference lies in temporal symmetry, which can avoid the model overfitting time order in the binary change branch, whereas other existing models cannot guarantee this point.

\section{Experiments}
\label{sec:exp}

The experiments are conducted on eight public remote sensing change detection datasets with diverse domains, multi-class geospatial objects, and multiple change types for comprehensive analysis.
xView2 \citep{gupta2019creating}, LEVIR-CD \citep{chen2020spatial}, WHU-CD \citep{ji2018fully}, S2Looking \citep{shen2021s2looking} and SpaceNet8 \citep{hansch2022spacenet} are used for single-temporal supervised object change detection.
CDD \citep{lebedev2018change} is used for bitemporal supervised binary change detection.
S2Looking \citep{shen2021s2looking} is also used for bitemporal supervised object change detection.
DynamicEarthNet \citep{toker2022dynamicearthnet} is used for single-temporal supervised and bitemporal supervised semantic change detection.
SECOND \citep{yang2021asymmetric} is only used for bitemporal supervised semantic change detection because of its incomplete land-cover annotation.

\subsection{\small Single-Temporal Supervised Object Change Detection}
We conduct comprehensive experiments of single-temporal supervised object change detection in multiple scenarios, i.e., in-domain building damage change, cross-domain generic building change, and in-domain road change due to flooding, to sufficiently evaluate the generalization of the proposed methods.

\bfsec{Training Datasets.}
Three HSR remote sensing segmentation datasets were used to train segmentation models and object change detectors by single-temporal supervision.
\begin{itemize}
\item \textbf{xView2 pre-disaster.} We used a subset of the xView2 dataset \citep{gupta2019creating}, namely xView2 pre-disaster, which is made up of the pre-disaster images and their annotations from \texttt{train} set and \texttt{tier3} set.
The xView2 pre-disaster dataset consists of 9,168 HSR optical remote sensing images with a total of 316,114 building instances annotations in the context of the sudden-onset natural disaster.
The images were collected from the Maxar / DigitalGlobe Open Data Program, and each image has a spatial size of 1,024$\times$1,024 pixels.
\item \textbf{SpaceNet8 \citep{hansch2022spacenet} pre-event.} The pre-event part of SpaceNet8 is used for single-temporal road change detection.
SpaceNet8 dataset is used in the context of flooding disasters to detect flooded buildings and roads via pre-event and post-event image pairs, i.e., bitemporal building and road change detection.
We only use the annotations of pre-event road segmentation to train DPCC, ChangeStar, and ChangeStar2 models.
We randomly selected 80\% of SpaceNet8 pre-event as \texttt{train} set, and the remaining 20\% is used as \texttt{val} set.
SpaceNet8 pre-event comprises 636 high-resolution optical images with RGB bands and sub-meter spatial resolutions.
Each image has a fixed spatial size of 1,300$\times$ 1,300 pixels.
\item \textbf{WHU-CD \citep{ji2018fully}.} This dataset is used for in-domain generalization evaluation. Each image has a spatial size of 15,354$\times$32,507 pixels with a spatial resolution of 0.2 m.
We cropped these two images into non-overlapped patch pairs with a fixed size of 1,024$\times$1,024 pixels and then split them into 80\%/20\% subsets for training and validation.
For single-temporal supervised learning, we unpair the training set, resulting in a single-temporal dataset.
For bitemporal supervised oracle, we do not change the original paired training set.
\end{itemize}

\begin{table*}[htb]
   \caption{\textbf{Single-temporal Supervised Object Change Detection} benchmark results on \textbf{xView2 building damage}, \textbf{LEVIR-CD}$^{\texttt{all}}$, \textbf{S2Looking}, \textbf{SpaceNet8 Flooded Road}, and \textbf{WHU-CD} datasets.
      The backbone network of all models is ResNet-50, pre-trained on ImageNet.
      All methods were trained using only single-temporal images and their semantic segmentation labels, except oracles.
      \label{tab:benchmark}}
   \centering
   \small
   \tablestyle{3pt}{1.5}
   \resizebox{\linewidth}{!}{
   \begin{tabular}{l|l|ll|ll|ll|ll|ll}
      \multirow{3}{*}{Method} & \multirow{3}{*}{Dense Feature} & \multicolumn{6}{c|}{\textbf{Train on} xView2 pre-disaster} & \multicolumn{2}{c|}{\textbf{Train on} SN8 pre-event}  &    \multicolumn{2}{c}{\textbf{Train on} WHU-Building}  \\ \cline{3-12}
          &  & \multicolumn{2}{c|}{xView2 building damage} & \multicolumn{2}{c|}{LEVIR-CD$^{\texttt{all}}$}   & \multicolumn{2}{c|}{S2Looking} & \multicolumn{2}{c|}{SN8 Flooded Road}   & \multicolumn{2}{c}{WHU-CD}   \\ \cline{3-12}
             &        & IoU (\%)    & F$_1$ (\%)   & IoU (\%) & F$_1$ (\%) & IoU (\%) & F$_1$ (\%) & IoU (\%) & F$_1$ (\%) & IoU (\%) & F$_1$ (\%) \\ \hline
      DPCC                    & PSPNet \citep{zhao2017pyramid}        & 14.4  & 25.1   & 55.9     & 71.7  & 9.4 & 17.2 &  31.1 &  47.4   & 44.6 & 61.7 \\
      ChangeStar              & + ChangeMixin                        & 19.4  & 32.5   & 61.6     & 76.3  &11.7 & 20.9 &  32.4   &  48.9 & 69.1 & 81.7   \\
      ChangeStar2             & + ChangeMixin2                       & 20.4\up{1.0} & 33.8\up{1.3} & 68.9\up{7.3} & 81.6\up{5.3} &18.8\up{7.1} &31.6\up{10.7} & 33.2\up{0.8}&49.9\up{1.0}& 70.9\up{1.8} & 83.0\up{1.3} \\ 
      ChangeStar2             & + ChangeMixin2 & \multicolumn{8}{c|}{Bitemporal Supervised Learning on WHU-CD (oracle)$\rightarrow$} & 72.5 &  84.1  \\
      \hline
      DPCC                    & DeepLab v3 \citep{chen2017rethinking} & 14.2 &  24.9  & 54.8     & 70.8  & 8.8  & 16.2 & 33.5  & 50.2  & 45.4 & 62.4\\
      ChangeStar              & + ChangeMixin                        & 19.0 & 32.0   & 61.0     & 75.7  & 11.4 & 20.5 &  34.9  & 51.7  & 73.2 & 84.6 \\
      ChangeStar2             & + ChangeMixin2                       & 19.7\up{0.7}  & 32.9\up{0.9} & 71.2\up{10.2}&83.2\up{7.5}&17.1\up{5.7}&29.1\up{8.6}& 35.1\up{0.2}&52.0\up{0.3} & 74.3\up{1.1}& 85.3\up{0.7} \\ 
      ChangeStar2             & + ChangeMixin2 & \multicolumn{8}{c|}{Bitemporal Supervised Learning on WHU-CD (oracle)$\rightarrow$} & 71.1 &  83.1  \\
      \hline
      DPCC                    & FarSeg \citep{zheng2020foreground}    & 14.5  & 25.4  & 55.1     & 71.0 & 9.9  & 17.9  &  32.7   & 49.3   & 44.9 & 62.0    \\
      ChangeStar              & + ChangeMixin                        & 19.5 & 32.7   & 65.7     & 79.3 & 11.9 & 21.2  &  35.6   & 52.5   & 71.4 & 83.3   \\
      ChangeStar2             & + ChangeMixin2                       & 21.8\up{2.3} & 35.7\up{3.0} &70.5\up{4.8} &82.7\up{3.4}  &19.5\up{7.6}&32.7\up{11.5}& 36.2\up{0.6} &  53.2\up{0.7} & 72.8\up{1.4} & 84.3\up{1.0}   \\
      ChangeStar2             & + ChangeMixin2 & \multicolumn{8}{c|}{Bitemporal Supervised Learning on WHU-CD (oracle)$\rightarrow$} & 74.4 &  85.3  \\
   \end{tabular}
   }
\end{table*}
\begin{figure*}[!htb]
   \centering
   \includegraphics[width=.9\linewidth]{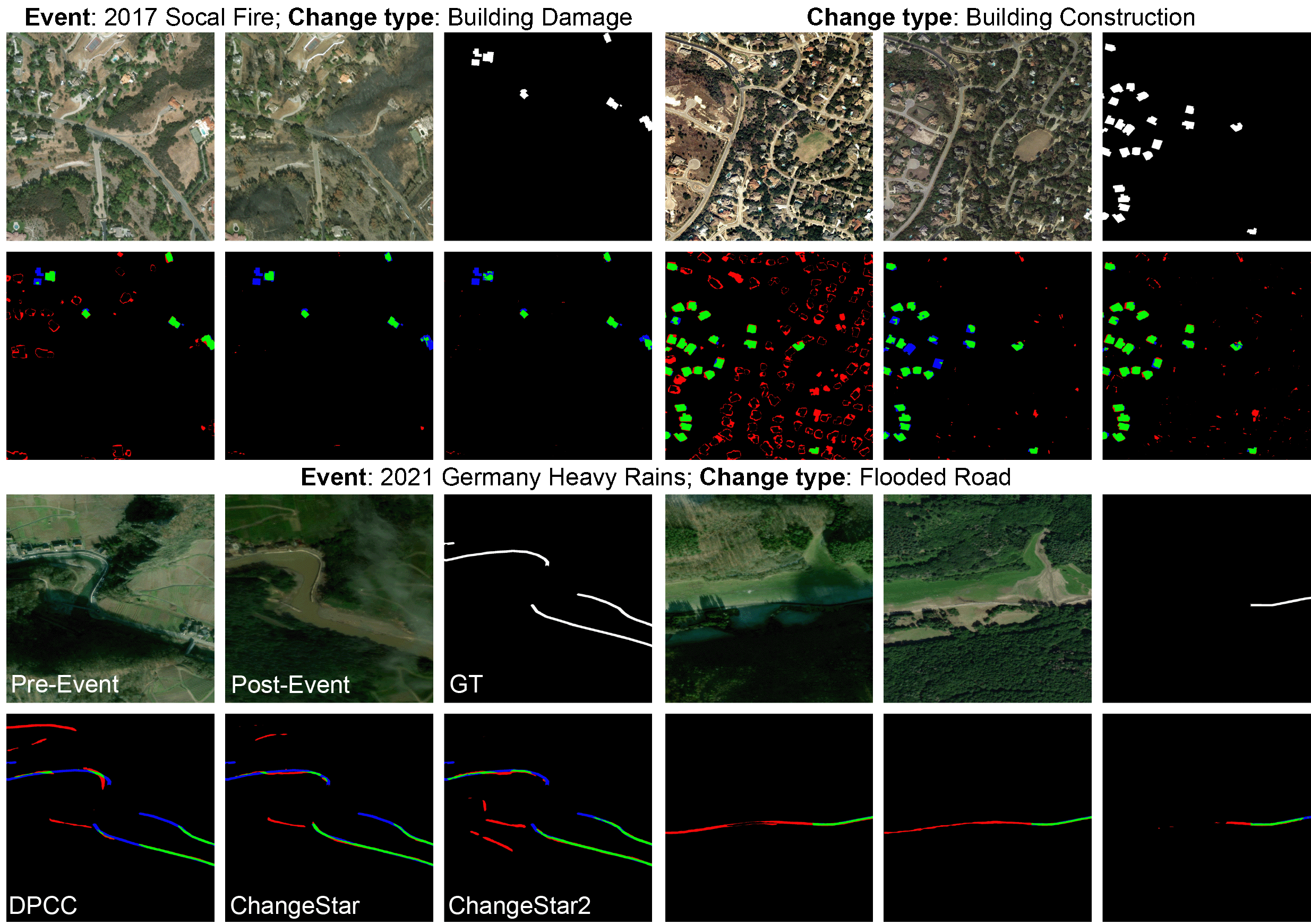}
   \caption{\textbf{Single-temporal Supervised Object Change Detection} visualization results of DPCC, ChangeStar and ChangeStar2 on xView2 building damage, LEVIR-CD, and SpaceNet8 Flooded Road datasets.
   All models are FarSeg-based variants.
   The rendered colors represent \textcolor[rgb]{0,1,0}{true positives (TP)}, \textcolor[rgb]{1,0,0}{false positives (FP)}, and \textcolor[rgb]{0,0,1}{false negatives (FN)}.
   These four sub-figures have a common layout that is annotated by white text shown in the sub-figure on the bottom left. 
   }
   \label{fig:xView2_LEVIR_SN8_viz}
\end{figure*}

\bfsec{Evaluation Datasets.}
We conduct zero-shot generalization evaluation for object change detection using two large-scale remote sensing building change detection datasets.

\begin{itemize}
   \item \textbf{xView2 building damage.} The \texttt{hold} set of the xView2 dataset is used for evaluation, which consists of 933 bitemporal image pairs.
   The xView2 dataset is originally used for multi-class change detection, i.e., building damage assessment.
   In this setting, we simplified it as a binary object change detection by only determining whether this building is damaged (i.e., \texttt{Major damage} and \texttt{Destroyed}).
   \item \textbf{LEVIR-CD.} LEVIR-CD dataset \citep{chen2020spatial} consists of 637 bitemporal remote sensing image pairs, which were collected from the Google Earth platform.
   Each image has a spatial size of 1,024$\times$1,024 pixels with a spatial resolution of 0.5 m.
   For annotation, this dataset provides a total of 31,333 change labels of building instances but without semantic labels.
   This dataset includes not only buildings appearing but also buildings disappearing for more general building changes.
   LEVIR-CD dataset is officially split into \texttt{train}, \texttt{val}, and \texttt{test}, three parts of which include 445, 64, and 128 pairs, respectively.
   If not specified, the entire dataset (LEVIR-CD$^{\texttt{all}}$) is used for evaluation.
   \item \textbf{S2Looking \citep{shen2021s2looking}.} This dataset features an off-nadir building change detection problem to evaluate robustness and generalization.
   \texttt{test} split is used to report performance.
   More details of this dataset can be seen in Sec~\ref{sec:bsocd}.
   \item \textbf{SpaceNet8 Flooded Road.} The flooded road part of SpaceNet8 is used to evaluate road change detection.
   The remaining 20\% SpaceNet8 pre-event and its corresponding post-event part are used as \texttt{val} set, which consists of 163 bitemporal image pairs and their flooded road map as the ground truth of road change map.
   \item \textbf{WHU-CD.} The remaining 20\% image pairs are used as the validation set to evaluate in-domain performance for single-temporal and bitemporal supervised learning simultaneously.
\end{itemize}

\bfsec{Implementation detail.}
Unless otherwise specified, all models were trained for 40k iterations with a \texttt{poly} learning rate policy, where the initial learning rate was set to 0.03 and multiplied by $(1 - \frac{\texttt{step}}{\texttt{max\_step}})^{\gamma}$ with $\gamma = 0.9$.
We used SGD as the optimizer with a mini-batch of 16 images, weight decay of 0.0001, and momentum of 0.9.
For training data augmentation, after the horizontal and vertical flip, rotation of $90\cdot k~(k=1,2,3)$ degree, and scale jitter, the images are then randomly cropped into 512$\times$512 pixels for xView2 pre-disaster dataset and 1,024$\times$1,024 pixels for SpaceNet8 dataset.

\bfsec{Metrics.}
The binary object change detection belongs to the pixel-wise binary classification task, therefore, we adopt intersection over union (IoU) and F$_1$ score of the foreground class to evaluate the object change detection.

\bfsec{Results.}
When only single-temporal supervision is available, DPCC families based on strong semantic segmentation models are the only available and reasonable baselines.
DPCC can be seen as an enhanced PCC method that compares classification maps from deep semantic segmentation models to achieve bitemporal remote sensing change detection in a general context.
Thus, we compare ChangeStar families against DPCCs with many representative deep segmentation models \citep{zhao2017pyramid,chen2017rethinking,zheng2020foreground}.
The quantitative comparison presented in Table~\ref{tab:benchmark} shows that ChangeStar significantly outperforms DPCC with three different segmentation models in three different typical scenarios.
Notably, these improvements only come at the cost of a slight overhead, which confirms the feasibility and significance of learning change representation via single-temporal supervision.
Besides, ChangeStar2 further consistently improves performances compared to ChangeStar.
From the qualitative results shown in Fig.~\ref{fig:xView2_LEVIR_SN8_viz}, we find that DPCC families mainly fail in false positives since they cannot learn temporal correlation to handle geometric offset and radiation difference.
ChangeStar families can significantly reduce these false positives by learning change representation from position-independent semantic differences.
ChangeStar2 families further sweep the remaining false positives by alleviating the absence of negative examples (unchanged object regions) from the perspective of learning better change representation.
For in-domain generalization evaluation, we trained ChangeStar2 via bitemporal supervised learning on WHU-CD as oracles.
We can find that the accuracy of ChangeStar2 families is close to the oracle, even if the variant based on DeepLab v3 exceeds its oracle. 
This suggests that single-temporal supervised learning is promising for in-domain generalization.
To evaluate out-of-domain generalization in the scenario with large variation in off-nadir angle, we test single-temporal methods on S2Looking.
ChangeStar2 also consistently achieves significant improvement over ChangeStar. 
However, its absolute accuracies are lower than that of in-domain scenarios (WHU-CD) and other out-of-domain scenarios with smaller domain gaps (LEVIR-CD).
This is an open problem for single-temporal supervised learning.
We hope this out-of-domain generalization can be improved in future works.


\begin{table*}[!htb]
   \caption{\textbf{Semantic Change Detection} benchmark results on \textbf{DynamicEarthNet} \texttt{val*}.
   The backbone networks of all entries are EfficientNet-B0 pretrained on ImageNet for a fair comparison.
   \label{tab:DEN_scd}}
   \centering
   \small
   \tablestyle{2pt}{1.5}
   \resizebox{\linewidth}{!}{
      \begin{tabular}{l|llll|cccccc|cccccc}
         \multirow{2}{*}{Method}                                 & \multirow{2}{*}{SCS ($\uparrow$)} & \multirow{2}{*}{BC ($\uparrow$)} & \multirow{2}{*}{SC ($\uparrow$)} & \multirow{2}{*}{mIoU ($\uparrow$)} & \multicolumn{6}{c|}{SC per class ($\uparrow$)} & \multicolumn{6}{c}{Seg. IoU per class ($\uparrow$)}                                                                                             \\ \cline{6-17}
                                                                 &                                   &                                  &                                  &                                    & Imp. Surf.                                     & Agric.                                              & Forest & Wetlands & Soil & Water & Imp. Surf. & Agric. & Forest & Wetlands & Soil & Water \\ \shline
         \multicolumn{2}{l}{\textit{Bitemporal Supervised}}      &                                   &                                  &                                  &                                    &                                                &                                                     &        &          &      &       &            &        &        &                         \\
         HRSCD str4 \citep{daudt2019multitask}                    & 18.9                              & 12.0                             & 25.9                             & 41.4                               & 15.7                                           & 4.6                                                 & 52.6   & 0.0      & 37.0 & 45.4  & 22.6       & 10.5   & 70.9   & 0.0      & 55.4 & 89.0  \\
         ChangeMask \citep{zheng2022changemask}                   & 19.5                              & 12.8                             & 26.2                             & 39.2                               & 14.6                                           & 3.9                                                 & 52.2   & 0.0      & 45.2 & 41.3  & 19.6       & 8.9    & 71.6   & 0.0      & 57.5 & 77.5  \\
         ChangeStar2 (U-Net)  & 19.8& 12.9& 26.7& 42.6 & 13.1& 7.4& 51.2& 0.0& 41.8& 46.9 & 23.0& 6.9& 73.4& 0.0& 60.6& 91.4  \\
         ChangeStar2 (FarSeg) & 19.4 & 12.9 & 26.0 & 41.8 & 10.7& 5.3& 50.3& 0.0& 42.5& 47.0 & 17.8& 7.9& 71.3& 0.0& 60.6& 93.0 \\
         \hline
         \multicolumn{2}{l}{\textit{Single-Temporal Supervised}} &                                   &                                  &                                  &                                    &                                                &                                                     &        &          &      &       &            &        &        &          &              \\
         DPCC (U-Net) \citep{ronneberger2015u}                    & 18.4                              & 11.1                             & 25.8                             & 42.5                               & 16.9                                           & 8.8                                                 & 49.0   & 0.0      & 38.9 & 41.3  & 22.7       & 9.5    & 73.6   & 0.0      & 58.8 & 90.0  \\
         ChangeStar2 (U-Net)                                     & 19.9\up{1.5}                      & 12.9\up{1.8}                     & 26.9\up{1.1}                     & 42.8\up{0.3}                       & 18.7                                           & 7.6                                                 & 49.8   & 0.0      & 36.6 & 48.6  & 22.7       & 10.4   & 72.2   & 0.0      & 58.2 & 93.2  \\ \hline
         DPCC (FarSeg) \citep{zheng2020foreground} & 19.0& 13.0 & 25.1 & 41.3  & 10.9   & 2.7  & 50.6   & 0.0    & 40.6 & 45.5  & 21.8& 7.7 & 69.8& 0.0  & 55.9 & 92.6  \\
         ChangeStar2 (FarSeg)  & 20.2\up{1.2} & 15.1\up{2.1} & 25.3\up{0.2} & 42.1\up{0.8}  & 14.6& 9.1& 49.0& 0.0& 36.7& 42.7 & 19.9& 10.2& 72.8& 0.0& 57.6& 92.4  \\
      \end{tabular}
   }
\end{table*}
\begin{figure*}[!htb]
   \centering
   \includegraphics[width=.98\linewidth]{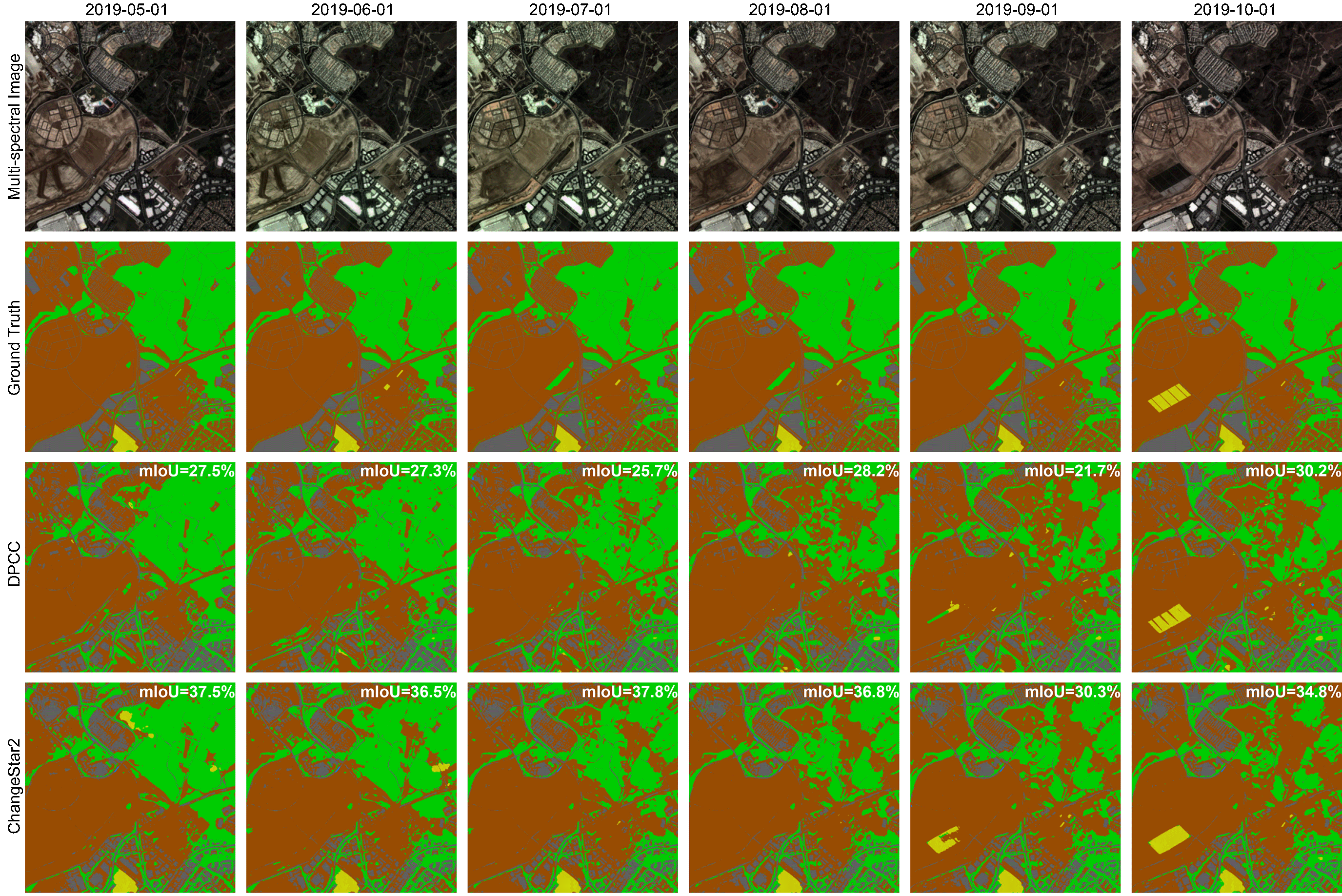}
   \caption{\textbf{Single-temporal Supervised Semantic Change Detection} visualization results of DPCC and ChangeStar2 on DynamicEarthNet \texttt{val*}.
   These two models are FarSeg-based variants.
   Legend: \clegend{ImpSurface}{Impervious Surface}, \clegend{Agriculture}{Agriculture}, \clegend{Forest}{Forest}, \clegend{Wetlands}{Wetlands}, \clegend{Soil}{Soil}, \clegend{Water}{Water}
   }
   \label{fig:den_viz}
\end{figure*}

\subsection{Single-Temporal Supervised Semantic Change Detection}
In this section, we further demonstrate the capability of STAR for semantic change detection in multi-spectral time-series satellite images.

\bfsec{Dataset.} DynamicEarthNet \citep{toker2022dynamicearthnet} is a time-series multi-spectral satellite dataset for land-cover and land-use (LULC) change detection, where the multi-spectral time-series satellite images were collected from Planet with a spatial resolution of 3 m and four bands of RGB and near-infrared.
It also provides monthly dense semantic labels of seven LULC classes, i.e., impervious surfaces (Imp. Surf.), agriculture (Agric.), forest \& other vegetation (Forest), wetlands, soil, water, and snow \& ice.
In this experiment, we focus on detecting semantic (multi-class) changes in time-series multi-spectral satellite images using ChangeStar2 with generalized STAR.
Thus, we reorganize the data of the released 55 AOIs for bitemporal-supervised or single-temporal supervised training and standard time-series validation.
We randomly select 11 AOIs from the released 55 AOIs as local validation set (DynamicEarthNet \texttt{val*}\footnote{\url{https://z-zheng.github.io/assets/changestar2/DEN/val\_star.txt}}), and the rest 44 AOIs are used as the training set.
To only provide single-temporal supervision, we separate each bitemporal image pair of the training set and globally shuffle the order.

\bfsec{Implementation detail.}
All models were trained for 20k iterations with a \texttt{poly} learning rate policy, where the initial learning rate was set to 0.03 and multiplied by $(1 - \frac{\texttt{step}}{\texttt{max\_step}})^{\gamma}$ with $\gamma = 0.9$.
We used SGD as the optimizer with a mini-batch of 16 images, weight decay of 0.0001, and momentum of 0.9.
For training data augmentation, after the horizontal and vertical flip, rotation of $90\cdot k~(k=1,2,3)$ degree, color jitter, and randomly 512$\times$512 cropping are used.

\bfsec{Metrics.}
Three official metrics \citep{toker2022dynamicearthnet}, i.e., binary change (BC) score, semantic change (SC) score, and semantic change segmentation (SCS) score are adopted for accuracy assessment.
SCS score is the average of BC score and SC score to reflect the overall performance.
We follow the official evaluation protocol to compute these metrics on all adjacent frames.

\bfsec{Results.}
We compare ChangeStar2 against the official baseline method \citep{toker2022dynamicearthnet}, i.e., U-Net-based DPCC, and our enhanced FarSeg-based DPCC as an extra stronger baseline.
As the quantitative comparison presented in Table~\ref{tab:DEN_scd}, U-Net-based and FarSeg-based ChangeStar2 consistently outperforms DPCC from three perspectives: semantic change detection, binary change detection, and land-cover segmentation.
It suggests that position-independent semantic difference information not only helps change detection but also time-series land-cover segmentation since learning representation from semantic difference can directly enhance inter-class discrimination of features.
The qualitative results are shown in Fig.~\ref{fig:den_viz}, which presents the superiority of ChangeStar2 in capturing subtle temporal dynamics.

\begin{table}[htb]
   \caption{\textbf{Bitemporal Supervised Binary Change Detection} benchmark results on \textbf{CDD} \texttt{test}.
   The amount of floating point operations (Flops) is computed with a float32 tensor of shape [2, 256, 256, 3] as input.
   ChangeStar2 families achieve top results.
      \label{tab:cdd}}
   \centering
   \small
   \tablestyle{1.5pt}{1.5}
   \resizebox{\linewidth}{!}{
   \begin{tabular}{l|c|ccc|cc}
      Method                             & backbone        & F$_1$         & Prec. & Rec. & \#Params. & \#Flops \\
      \shline
      \cb\textit{ConvNet-based}             &                &                &       &      &            &         \\
      FC-Siam-Diff \citep{daudt2018fully} & -               & 65.2          & 76.2  & 57.3 & 1.3M       & 4.7G     \\
      ESCNet \citep{zhang2021escnet}      & -               & 93.5          & 90.9  & 96.4 & 5.3M       & -        \\ 
      \textbf{ChangeStar2}               & Sv2 0.5$\times$ & \textbf{95.7} & 96.1  & 95.5 & 1.4M       & 3.0G     \\
      \hline
      DSAMNet \citep{shi2021deeply}       & R-18          &   93.7        &  94.5  &  92.8  &   17.0M    & 75.3G   \\
      SNUNet-C32 \citep{fang2021snunet}   & -              &   95.3        & 95.6  &  94.9  &   12.0M    & 54.7G   \\
      ChangeStar \citep{zheng2021change}  & R-18            &  96.8         & 97.5  & 96.1  &  16.4M     & 16.3G   \\
      \textbf{ChangeStar2}               & R-18            & \textbf{96.9} & 97.4  & 96.3 & 16.9M       & 17.6G   \\
      \hline
      IFN \citep{zhang2020deeply}         & VGG-16          &   90.3        & 95.0  &  86.1  &   36.0M    & 82.2G \\
      SNUNet-C48 \citep{fang2021snunet}   & -               &   96.2        & 96.3  &  96.2  &   27.1M    & 123.0G   \\
      ChangeStar \citep{zheng2021change}  & R-50            & 97.5          & 98.1  & 97.0   & 31.1M      & 23.7G   \\
      \textbf{ChangeStar2}               & R-50            & \textbf{97.5} & 98.0  & 97.0   & 31.6M      & 24.9G  \\ 
      \hline
      \vb\textit{Transformer-based}         &                 &                &       &      &            &         \\
      BiT \citep{chen2021remote, wang2022empirical}         & {\scriptsize ViTAEv2-S}       &   97.0        &  -     &    -    &     19.6M      &   15.7G   \\
      \textbf{ChangeStar2}               & MiT-B1          & 97.4               &  97.9      & 97.0      &   18.3M        &   15.8G     \\
      \textbf{ChangeStar2} (d512)  & Swin-T          & \textbf{98.0} & \textbf{98.4} &  \textbf{97.6}    &   62.1M        &   96.5G     \\
   \end{tabular}
   }
\end{table}

\bfsec{Comparisons with bitemporal supervised methods.}
We benchmark two advanced bitemporal supervised semantic change detectors \citep{daudt2019multitask, zheng2022changemask}.
Besides, we also trained two variants of ChangeStar2 via bitemporal supervised learning to ablate the architecture improvement, thereby evaluating the effectiveness of single-temporal supervised learning.
These four methods are trained with bitemporal supervision from all adjacent frames in a time series.
The results are listed in Table~\ref{tab:DEN_scd}.
It can be seen that bitemporal supervised methods have on-par or better SCS scores compared to DPCC methods, mainly reflecting on generally better SC scores, which confirms the effectiveness of the bitemporal supervision from adjacent frames, as we all know.
It is surprising that ChangeStar2, as the single-temporal supervised method, outperforms these four bitemporal supervised methods in terms of overall performance, mainly achieving significantly better BC scores.
These results suggest that bitemporal supervision from arbitrary two frames is also effective and even better.
The underlying reason is that sufficient change supervisory signals can be guaranteed when using arbitrary two frames to construct a pseudo bitemporal image pair. 
In contrast, the changes between real-world adjacent frames in monthly sampled time series are very sparse and subtle, causing insufficient change supervisory signals.
It further confirms that STAR can be generalized to semantic change detection and works better in this time-series change detection.


\subsection{Bitemporal Supervised Binary Change Detection}
ChangeStar2 is a unified change detection architecture trained by generalized STAR as default, but it also can be trained by bitemporal supervised learning for bitemporal supervised change detection.
In this section, we demonstrate the effectiveness of ChangeStar2 architecture for bitemporal supervised binary change detection.
All compared models are trained by bitemporal supervised learning for a fair comparison.

\bfsec{Dataset.}
CDD dataset \citep{lebedev2018change} is used for bitemporal supervised binary change detection benchmark.
The original dataset consists of seven season-varying bitemporal image pairs with the size of 4,725$\times$2,700 pixels and four image pairs with the size of 1,900$\times$1,000 pixels, where each image was collected from Google Earth with spatial resolutions ranged from 0.03 m and 0.1 m.
Following the common practice \citep{zhang2021escnet, fang2021snunet, wang2022empirical}, the dataset is further regenerated by cropping into 256$\times$256 image patches, which results in a training set of 10,000 image pairs, a validation set of 3,000 image pairs, and a test set of 3,000 image pairs.
We trained all models on the training set and reported their results on the test set.

\bfsec{Metrics.}
F$_1$ score, precision rate (Prec.), and recall rate (Rec.) of change regions are used as evaluation metrics.

\bfsec{Implementation detail.}
Random flipping and rotations are used to augment training data.
SGD is used as the optimizer with a weight decay of 0.0001 and a momentum of 0.9.
Following common practice, we train for 200 epochs on \texttt{train} set with a total batch size of 16 
A ``poly'' learning rate policy ($\gamma$ = 0.9) is applied with an initial learning rate of 0.03.

\bfsec{Results.}
The quantitative results are presented in Table~\ref{tab:cdd}.
It is observed that ChangeStar2 families consistently achieve better performances in different configurations to meet various application scenarios.
Specifically, for lightweight ConvNet-based change detector, ChangeStar2 with shufflenet v2 0.5$\times$ achieves 95.7\% F$_1$ with only 1.4M parameters and 3.0 GFlops, which is even better than most existing heavy models with more powerful backbone (e.g., DSAMNet, IFN).
This result emphasizes that reusing an advanced segmentation model as a Siamese dense feature extractor is a more desirable way for change detection architecture design.
With R-18 or R-50 as the backbone, ChangeStar, and ChangeStar2 have similar performance and complexity.
They surpass existing methods (DSAMNet, SNUNet-C32, IFN, SNUNet-C48) by a large margin of F$_1$ and meanwhile with much less computational overhead.
For example, with R-18, ChangeStar2 outperforms DSAMNet by 3.2\% F$_1$ and only takes its 23\% Flops.
ChangeStar2 with R-50 outperforms SNUNet-C48 by 1.3\% and only takes its 20\% Flops. 
For advanced Transformer-based change detectors, BiT with ViTAEv2 is a SOTA method due to the large model capacity and long-range dependencies modeling of Transformers.
To compare with it, we build a Transformer-based variant of ChangeStar2 with MiT-B1 \citep{xie2021segformer} as the backbone to achieve highly similar parameters and Flops.
The result shows that ChangeStar2 outperforms BiT with ViTAEv2 by 0.4 F$_1$.
We further scale up ChangeStar2 to explore more potential. 
Benefiting from the idea of model reusing, we can easily adapt Swin-T \citep{liu2021swin} as the backbone and scale up the feature dimension of the FarSeg part from 256 to 512.
This new scaled-up variant, ChangeStar2 (d512) with Swin-T, achieves 98.0\% F$_1$, setting a new record on the CDD dataset.


\subsection{Bitemporal Supervised Object Change Detection}
\label{sec:bsocd}
In this section, we further demonstrate the effectiveness of ChangeStar2 architecture for bitemporal supervised object change detection.
Different from binary change detection, i.e., class-agnostic change detection, object change detection aims to detect the changes of the specific objects of interesting, which requires change detectors with stronger feature representation capability for the semantics of object and change.
All compared models are trained by bitemporal supervised learning for a fair comparison.

\bfsec{Dataset.}
S2Looking \citep{shen2021s2looking} is a larger scale, more complex, and challenging building change detection dataset due to tiny change objects, geometric offset, radiation difference, and diverse imaging angles.
The S2Looking dataset contains 5,000 image pairs with spatial resolutions from 0.5 to 0.8 m and 65,920 change instances.
The images were collected from GaoFen, SuperView, and BeiJing-2 satellites of China, which mainly covered globally distributed rural areas.
This dataset features side-looking satellite images, which pose a special yet important challenge that requires the change detector to have sufficient robustness to the registration error and the object geometric offset caused by off-nadir imaging angles.
Each image has a fixed size of 1,024$\times$1,024 pixels.
The official \texttt{train}/\texttt{val}/\texttt{test} split is used in all experiments.

\bfsec{Metrics.}
F$_1$ score, precision rate (Prec.), and recall rate (Rec.) of change regions are used as evaluation metrics.

\begin{table}[htb]
   \caption{\textbf{Bitemporal Supervised Object Change Detection} results on \textbf{S2Looking} \texttt{test}.
      The amount of floating point operations (Flops) is computed with a float32 tensor of shape [2, 512, 512, 3] as input.
      ChangeStar2 families achieve SOTA results.
      \label{tab:s2l}}
   \centering
   \small
   \tablestyle{1.5pt}{1.5}
   \resizebox{\linewidth}{!}{
   \begin{tabular}{l|c|ccc|cc}
      Method                             & backbone        & F$_1$         & Prec. & Rec. & \#Params. & \#Flops \\
      \shline
      \cb\textit{ConvNet-based}             &                &                &       &      &            &         \\
      FC-Siam-Diff \citep{daudt2018fully} & -               & 13.1          & 83.2  & 15.7 & 1.3M       & 18.7G   \\
      \textbf{ChangeStar2}               & Sv2 0.5$\times$ & 65.0          & 66.1  & 63.9 & 1.4M       & 10.9G   \\
      \hline
      STANet \citep{chen2020spatial}      & R-18            & 45.9          & 38.7  & 56.4 & 16.9M      & 156.7G  \\
      CDNet \citep{chen2021adversarial}   & R-18            & 60.5          & 67.4  & 54.9 & 14.3M      & -       \\
      BiT \citep{chen2021remote}          & R-18            & 61.8          & 72.6  & 53.8 & 11.9M      & 34.7G   \\
      ChangeStar \citep{zheng2021change}  & R-18            & 66.3          & 70.9  & 62.2 & 16.4M      & 65.3G   \\
      \textbf{ChangeStar2}               & R-18            & \textbf{67.4} & 68.5  & 66.3 & 16.4M      & 63.5G   \\
      \hline
      ChangeStar \citep{zheng2021change}  & R-50            & 67.4          & 70.8  & 64.3 & 31.0M      & 94.7G   \\
      \textbf{ChangeStar2}               & R-50            & \textbf{67.8} & 69.2  & 66.5 & 31.6M      & 97.9G  \\ \hline
      \vb\textit{Transformer-based}         &                 &                &       &      &            &         \\
      \textbf{ChangeStar2}               & MiT-B1          & 67.7          & 70.3  & 65.4 & 18.3M      & 66.5G  \\
      \textbf{ChangeStar2} (d512)        & Swin-T          & 68.8          & 69.8  & 67.8 & 62.1M      & 378.1G  \\
   \end{tabular}
   }
\end{table}

\bfsec{Implementation detail.}
Random flip, rotation, scale jitter, and cropping into 512$\times$512 are used as training data augmentation.
SGD is used as our optimizer, where the weight decay is 0.0001, and the momentum is 0.9.
The total batch size is 16, and an initial learning rate is 0.03.
We train for 60k iterations on \texttt{train} split.
A ``poly'' learning rate policy ($\gamma$ = 0.9) is applied, similar to \cite{zheng2021change}.
We train all models on \texttt{train} split and report results on \texttt{test} split.

\bfsec{Results.}
We compare ChangeStar2 to many representative and SOTA change detectors \citep{daudt2018fully, chen2020spatial, chen2021adversarial, chen2021remote, zheng2021change}
The results on S2Looking is in Table~\ref{tab:s2l}.
Without any architecture modification, ChangeStar2 still achieves superior performance both with R-18 and R-50, outperforming all existing methods by significant gaps.
Also, we adopt Transformers as the backbone to validate their compatibility with ChangeStar2 on the S2Looking dataset.
The results present consistent conclusions that ChangeStar2 with R-50 and with MiT-B1 achieve similar performances, and the scaled-up variant significantly improves the performance to 68.8\% F$_1$.
It suggests that ChangeStar2 has good scalability and compatibility with Transformers.

\begin{table*}[htb]
   \caption{\textbf{Bitemporal Supervised Semantic Change Detection} benchmark results on \textbf{SECOND} at the system level.
      The amount of Flops is computed with a float32 tensor of shape [2, 256, 256, 3] as input.
      \label{tab:second_benchmark}}
   \centering
   \small
   \tablestyle{5.5pt}{1.2}
   \resizebox{\linewidth}{!}{
   \begin{tabular}{l|cc|cc|c|cc}
      \multirow{2}{*}{Method}                                        & \multicolumn{2}{c|}{Binary Change} & \multicolumn{2}{c|}{Semantic Change} & \multirow{2}{*}{Overall (\%)$\uparrow$} & \multirow{2}{*}{\#Params} & \multirow{2}{*}{\#Flops$\downarrow$}                 \\
                                                                     & IoU (\%)                 & F$_1$ (\%)               & SeK (\%)$\uparrow$                            & Kappa (\%)                &                                      &       &       \\ \shline
      \multicolumn{1}{l|}{\cb\textit{ConvNet-based}}             &                          &                          &                                               &                           &                                      &       &       \\
      HRSCD strategy 4 \citep{daudt2019multitask}                     & 42.0                     & 59.1                     & 7.8                                           & 14.0                      & 23.5                                 & 2.3M  & 6.1G  \\
      Bi-SRNet (R-18-D8) \citep{ding2022bi}                                     & 52.6                     & 68.9                     & 16.1                                          & 25.9                      & 32.1                                 & 23.3M & 48.0G \\
      ChangeMask (EfficientNet-B0) \citep{zheng2022changemask}        & \textbf{54.2}                     & \textbf{70.3}                     & 17.8                                          & 28.2                      & 33.6                                 & 10.6M & 9.4G  \\
      \textbf{ChangeStar2} (EfficientNet-B0)                        &    53.1                   &  69.4    &                      \textbf{18.1}                    &      \textbf{29.0}      &    \textbf{33.7}     &   8.9M    &   12.2G     \\ 
      \hline
      \multicolumn{1}{l|}{\vb\textit{Transformer-based}}         &                          &                          &                                               &                           &                                      &       &       \\
      \textbf{ChangeStar2} (MiT-B1) &     52.5      &      68.8        &    18.5     &    29.8       &       33.9    &   18.3M    &   15.8G \\
      \textbf{ChangeStar2} (Swin-T, d512) &    53.8   &   70.0   &  19.2 &  30.5  &   34.6  &   62.1M    &  96.5G  \\ 
   \end{tabular}
   }
\end{table*}

\subsection{Bitemporal Supervised Semantic Change Detection}
In this section, we demonstrate the capability of ChangeStar2 architecture for semantic change detection without any architecture modification.
All compared models are trained by bitemporal supervised learning for a fair comparison.

\bfsec{Dataset.}
SECOND \citep{yang2021asymmetric} is a multi-city and multi-class change detection dataset, where 2,968 pairs of HSR remote sensing images with semantic annotations are publicly available.
These images were collected from several platforms and optical sensors.
Each image has a fixed size of 512$\times$512 pixels with the spatial resolution varying from 0.3 m to 5 m.
The category system of the SECOND dataset comprises six land-cover classes, i.e., non-vegetated ground surface, tree, low vegetation, water, buildings, and playgrounds.
Therefore, there are potentially 37 types of land-cover change that need to be detected.
We follow the dataset split (train\footnote{\url{https://z-zheng.github.io/assets/changemask/SECOND/train.csv}}/val\footnote{\url{https://z-zheng.github.io/assets/changemask/SECOND/val.csv}}) and training settings used in \cite{zheng2022changemask} for a fair comparison.

\bfsec{Metrics.}
Intersection over union (IoU) and F$_1$ of change regions are evaluation metrics for the binary change detection part.
Separated kappa coefficient (SeK) and vanilla kappa coefficient are evaluation metrics for semantic change detection.
SeK is computed by official tool\footnote{\url{http://www.captain-whu.com/project/SCD/}}.
Following \citep{zheng2022changemask}, a weighted score is adopted for judging overall performance, which is defined as $0.3 {\rm mIoU} + 0.7 {\rm SeK}$. 
Here, mIoU is the average of IoUs of change regions and unchanged regions.

\bfsec{Implementation detail.}
Random flipping, rotation, cropping into 256$\times$256, and color jitter were used as training data augmentation, following \citep{zheng2022changemask}.
We trained all models with a total batch size of 16 for 15k iterations.
SGD with a weight decay of 0.0001 and a momentum of 0.9 is used as the optimizer.
All backbone networks were pre-trained on ImageNet-1k.

\bfsec{Results.}
We compare ChangeStar2 with three representative methods \citep{daudt2019multitask, ding2022bi, zheng2022changemask} at the system level.
The quantitative results are presented in Table~\ref{tab:second_benchmark}.
With a ConvNet-based backbone, ChangeStar2 outperforms existing methods at the system level, suggesting its effectiveness in bitemporal supervised semantic change detection.
With the same backbone, ChangeStar2, as a unified change detector, is slightly better than ChangeMask, a specialized semantic change detector, in terms of overall performance.
It is observed that ChangeStar2 outperforms ChangeMask by 0.3\% SeK; however, in terms of binary change part, ChangeMask is much better than ChangeStar2.
From the perspective of multi-task feature representation, ChangeStar2 shares more feature maps in the encoder and the decoder, while ChangeMask only shares feature maps in the encoder.
Thus, ChangeStar2 has a more balanced performance on each sub-tasks.
To validate the compatibility with Transformers and scalability of ChangeStar2 for semantic change detection, we build two variants of ChangeStar2 with MiT-B0 or Swin-T as the backbone.
As expected, Transformer backbones further improve overall performances, and the improvement of scaling up ChangeStar2 is more significant, confirming the compatibility with the Transformer and scalability of ChangeStar2.

\subsection{Method Analysis}
To delve into the proposed method, we conducted comprehensive experiments using FarSeg-based variants with ResNet-50 if not specified since it is more robust than other variants.

\bfsec{Importance of Semantic Supervision.}
Semantic supervision not only enables ChangeStar to segment objects but also can facilitate object change representation learning.
Table~\ref{tab:as_component} (b)/(c) and (d)/(e) show that the introduction of semantic supervision is positive for object change detection.
Quantitatively, semantic supervision improves the baseline by 0.5\% IoU and 0.5\% F$_1$ and improves the baseline with temporal symmetry by 1.6\% IoU and 1.2\% F$_1$.
This indicates that semantic representation provided by semantic supervision facilitates object change representation learning, and object change representation is stronger when possessing temporal symmetry.

\bfsec{Importance of Temporal Symmetry.}
Temporal symmetry, as a mathematical property of binary change, can provide a prior regularization to learn a more robust change representation.
Table~\ref{tab:as_component} (b)/(d) and (c)/(e) shows that using temporal symmetry gives a 2.2\% IoU and 1.7\% F$_1$ gains over the baseline and gives a 3.3\% IoU and 2.4\% F$_1$ over the baseline with semantic supervision.
This indicates that it is significantly important to guarantee temporal symmetry in binary object change detection for STAR.
We can also find that temporal symmetry makes it better to learn change representation from semantic representation.

\begin{table}[htb]
   \caption{Single-temporal supervised object change detection results on LEVIR-CD$^{\texttt{all}}$ to understand the contribution of each component.
      \label{tab:as_component}}
   \centering
   \small
   \tablestyle{1.5pt}{1.5}
   \resizebox{\linewidth}{!}{
   \begin{tabular}{l|ccc|cc}
      Method         & STAR   & Semantic Sup. & Temporal Sym. & IoU (\%) & F$_1$ (\%) \\ \shline
      (a) DPCC   &        & \cmark        &               & 55.1    & 71.0      \\ \hline
      (b) Baseline   & \cmark &               &               & 61.9    & 76.4      \\
      (c) + Semantic & \cmark & \cmark        &               & 62.4    & 76.9      \\
      (d) + TSM      & \cmark &               & \cmark        & 64.1    & 78.1      \\
      (e) ChangeStar & \cmark & \cmark        & \cmark        & 65.7    & 79.3      \\
   \end{tabular}
   }
\end{table}

\begin{table}[htb]
   \caption{The accuracy of different label assignment strategies.
      \label{tab:label_assign}}
   \centering
   \small
   \tablestyle{17pt}{1.5}
   \begin{tabular}{l|cc}
      Method       & IoU (\%) & F$_1$ (\%) \\ \shline
      \texttt{or}  & 43.8     & 61.0       \\
      \texttt{xor} & 65.7     & 79.3       \\
   \end{tabular}
\end{table}

\bfsec{Label assignment.}
Here, we discuss the impact of different label assignment strategies on accuracy.
Table~\ref{tab:label_assign} presents that using logical \texttt{or} operation achieves 43.8\% IoU, while using logical \texttt{xor} operation achieves 65.7\% IoU.
This is because these negative samples (i.e., overlapped region) are necessary to make the model learn to suppress false positives that occurred on objects that have not changed in the period from $t_1$ to $t_2$, which can be satisfied by logical \texttt{xor}.
However, the logical \texttt{or} operation wrongly assigns their labels.

\begin{figure*}[htb]
   \centering
   \subfigure[$t_1$ Image]{
      \begin{minipage}[b]{0.15\linewidth}
         \includegraphics[width=\linewidth]{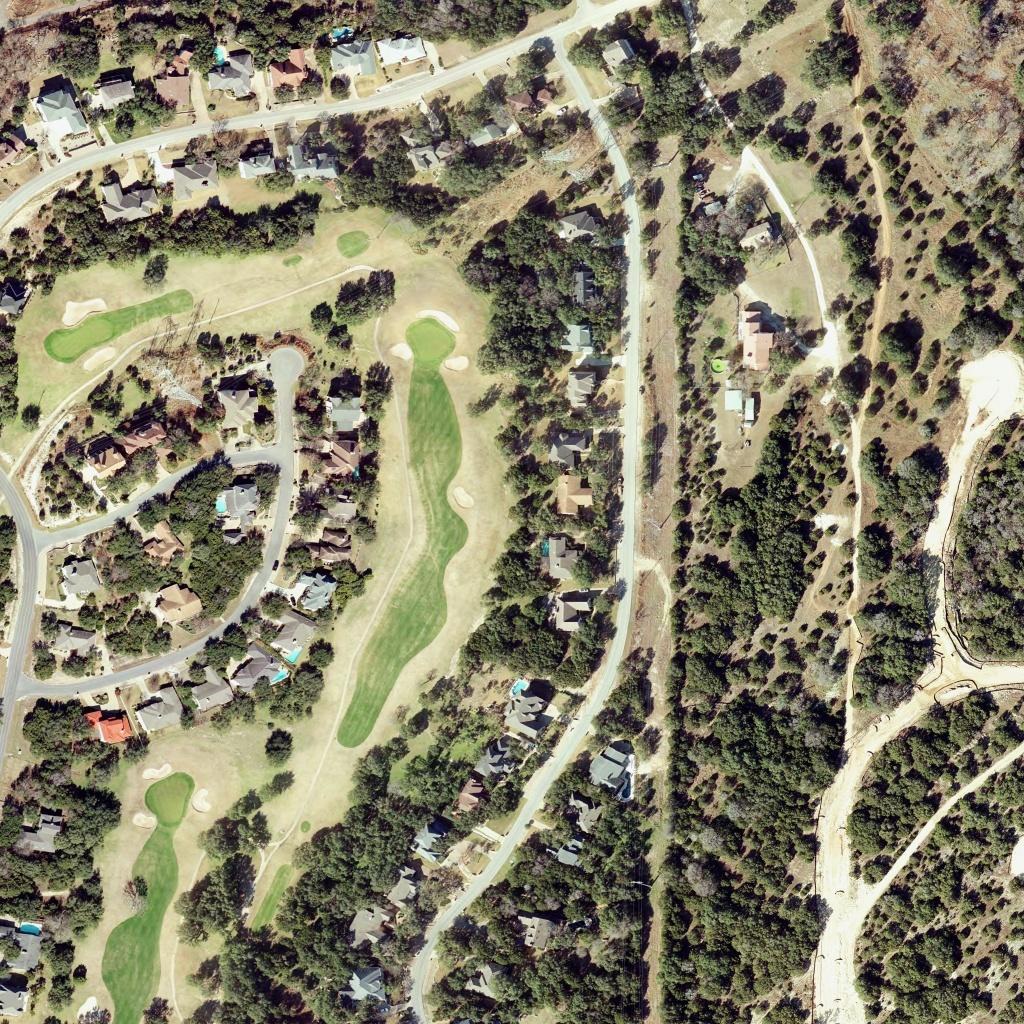}\vspace{4pt}
         \includegraphics[width=\linewidth]{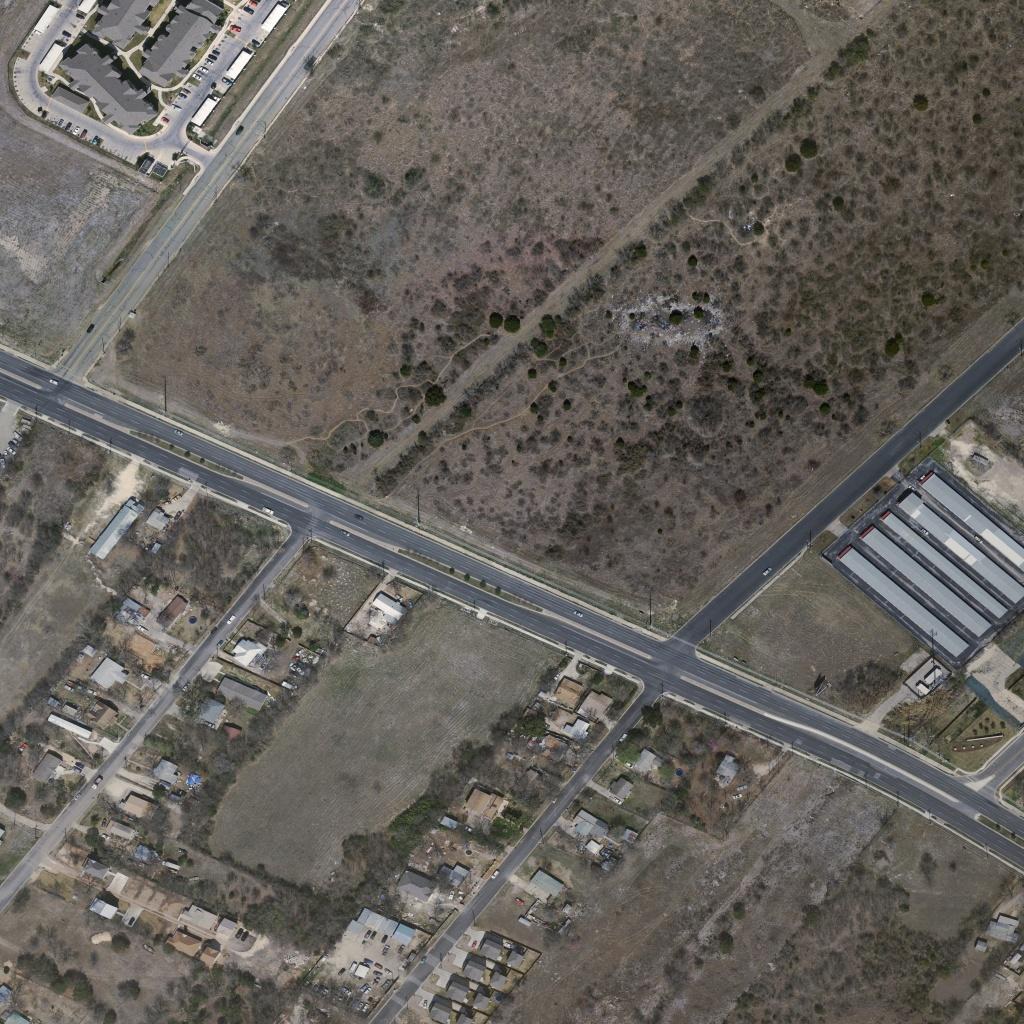}
      \end{minipage}
   }
   \subfigure[$t_2$ Image]{
      \begin{minipage}[b]{0.15\linewidth}
         \includegraphics[width=\linewidth]{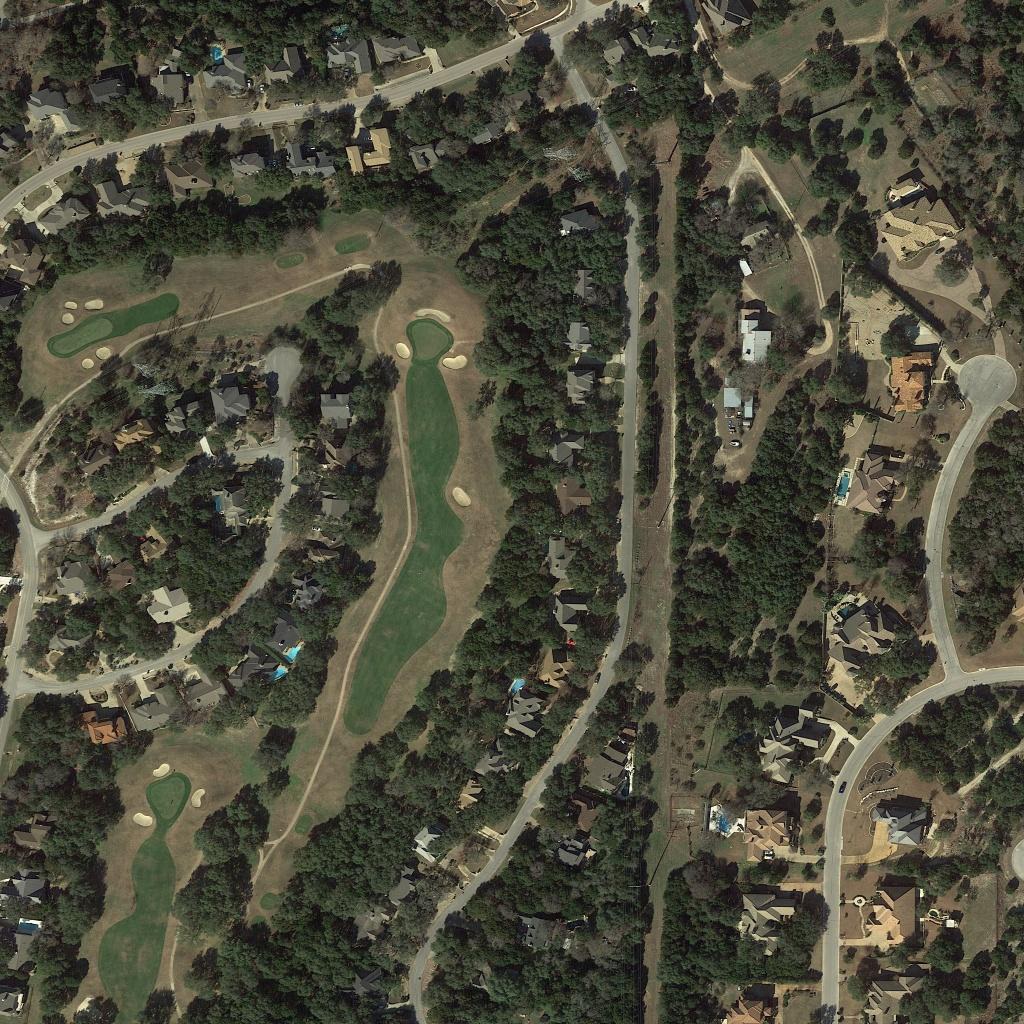}\vspace{4pt}
         \includegraphics[width=\linewidth]{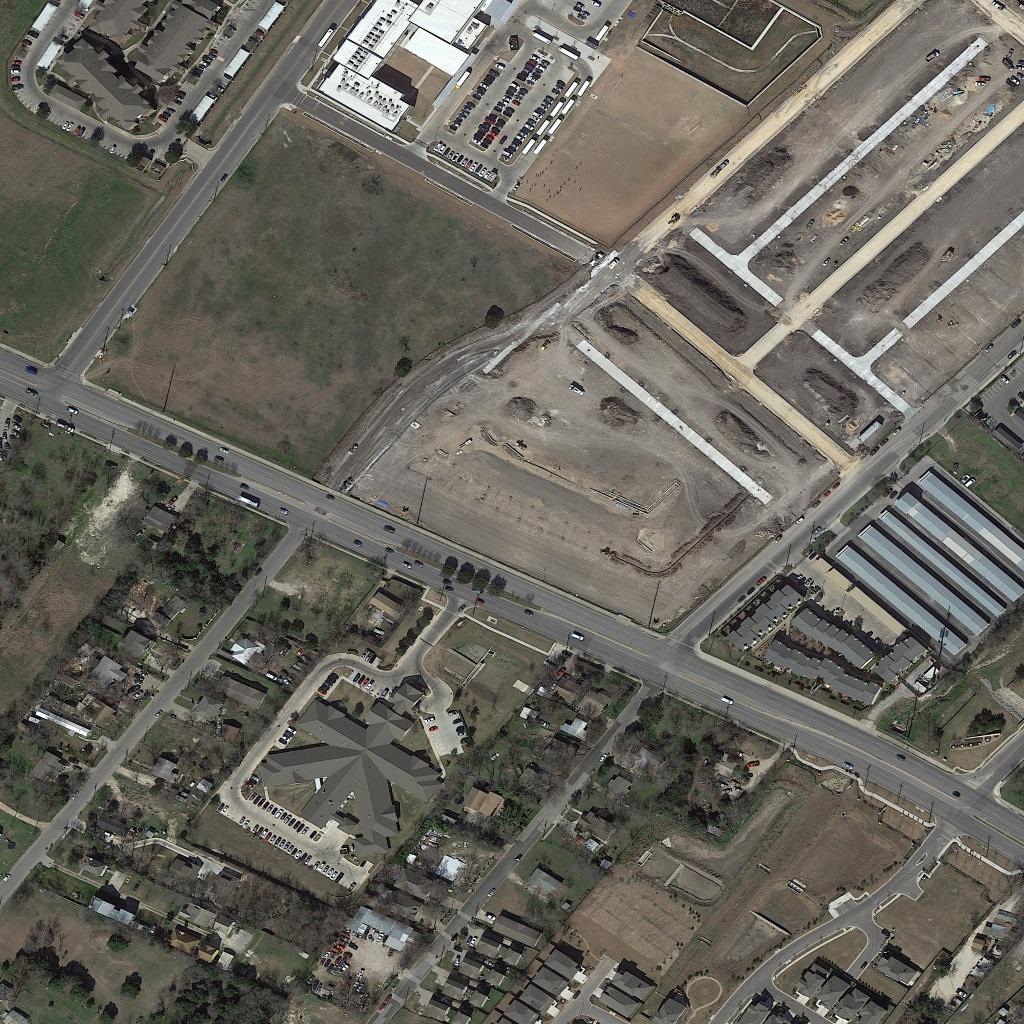}
      \end{minipage}
   }
   \subfigure[Ground Truth]{
      \begin{minipage}[b]{0.15\linewidth}
         \includegraphics[width=\linewidth]{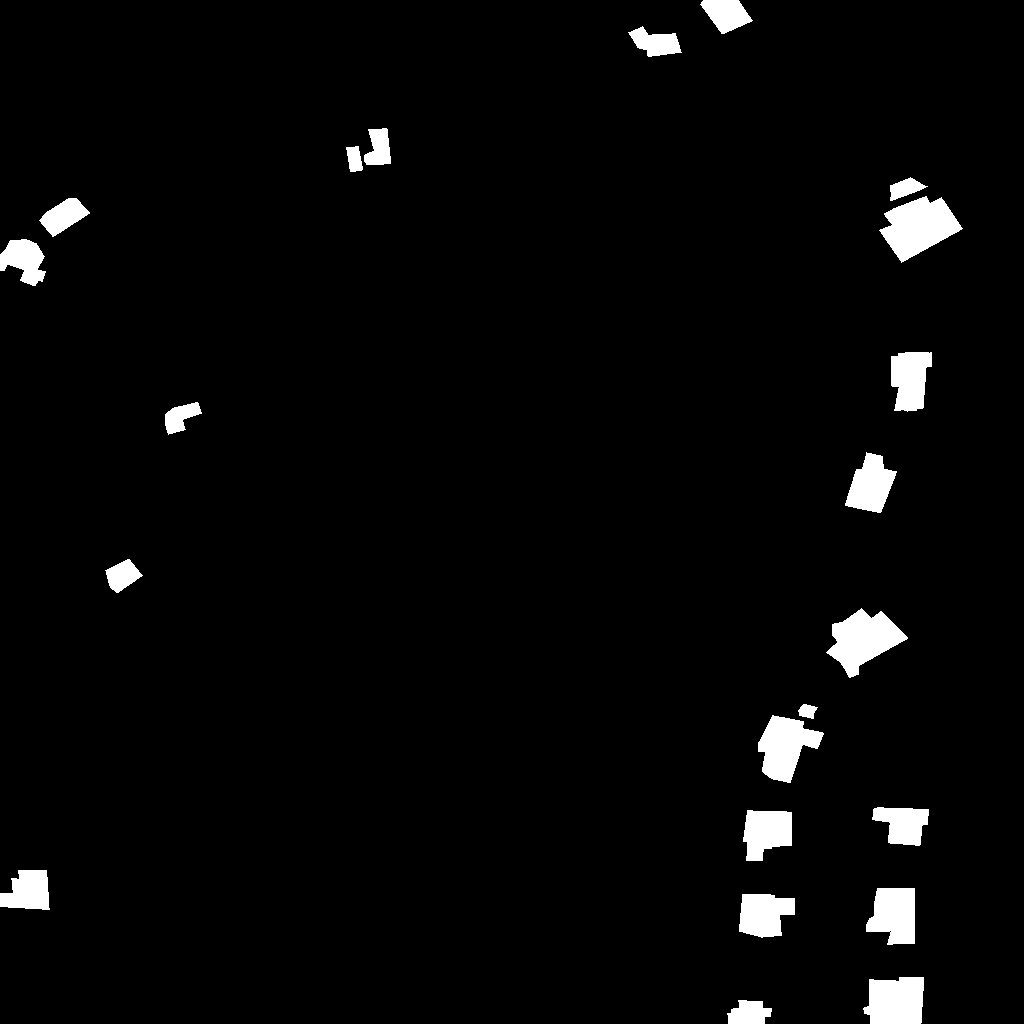}\vspace{4pt}
         \includegraphics[width=\linewidth]{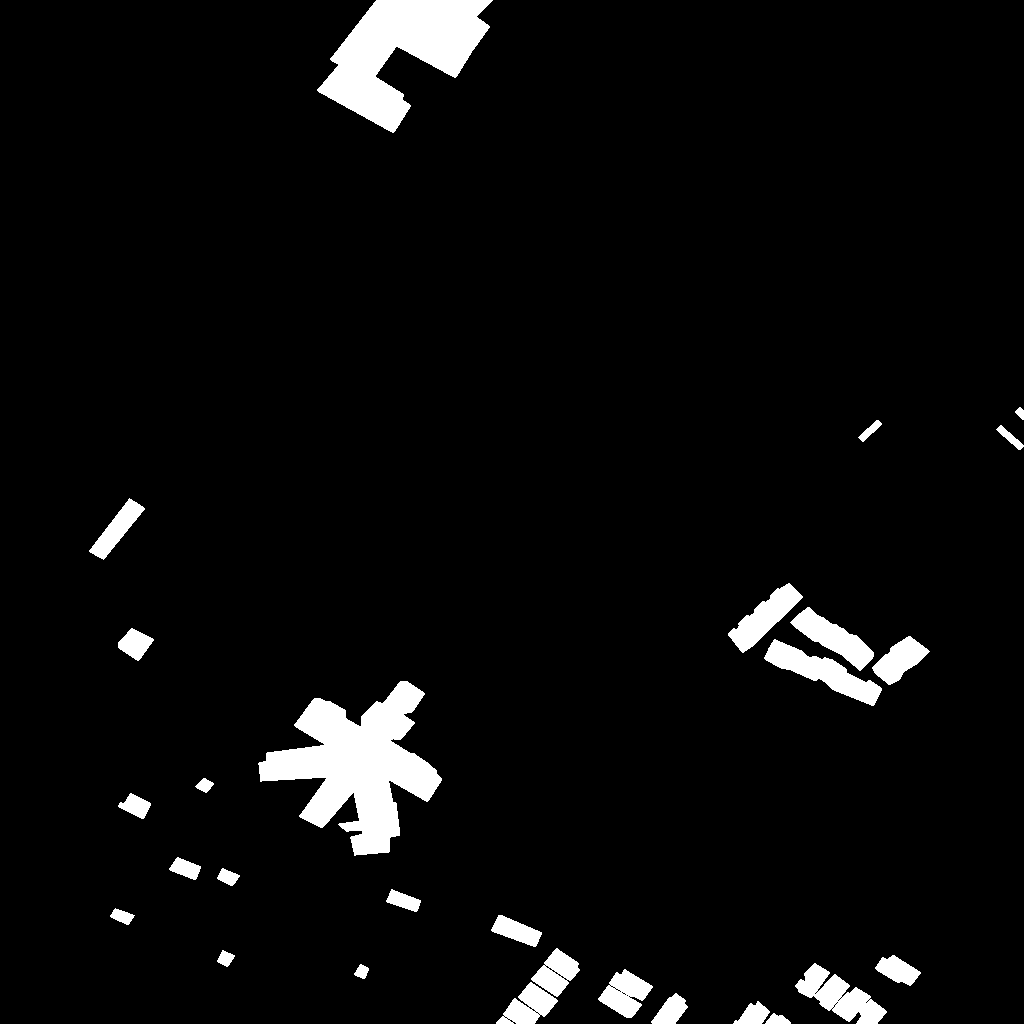}
      \end{minipage}
   }
   \subfigure[Bitemporal Sup.]{
      \begin{minipage}[b]{0.15\linewidth}
         \includegraphics[width=\linewidth]{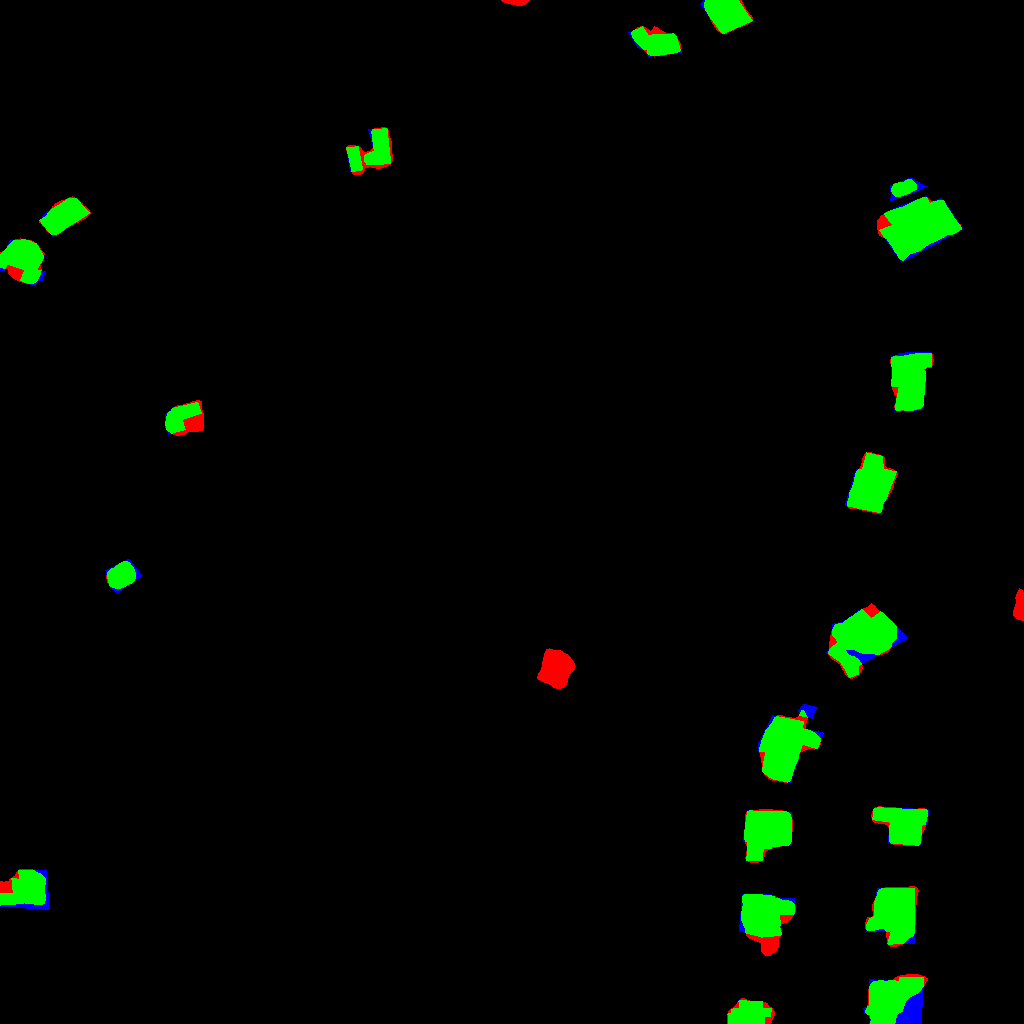}\vspace{4pt}
         \includegraphics[width=\linewidth]{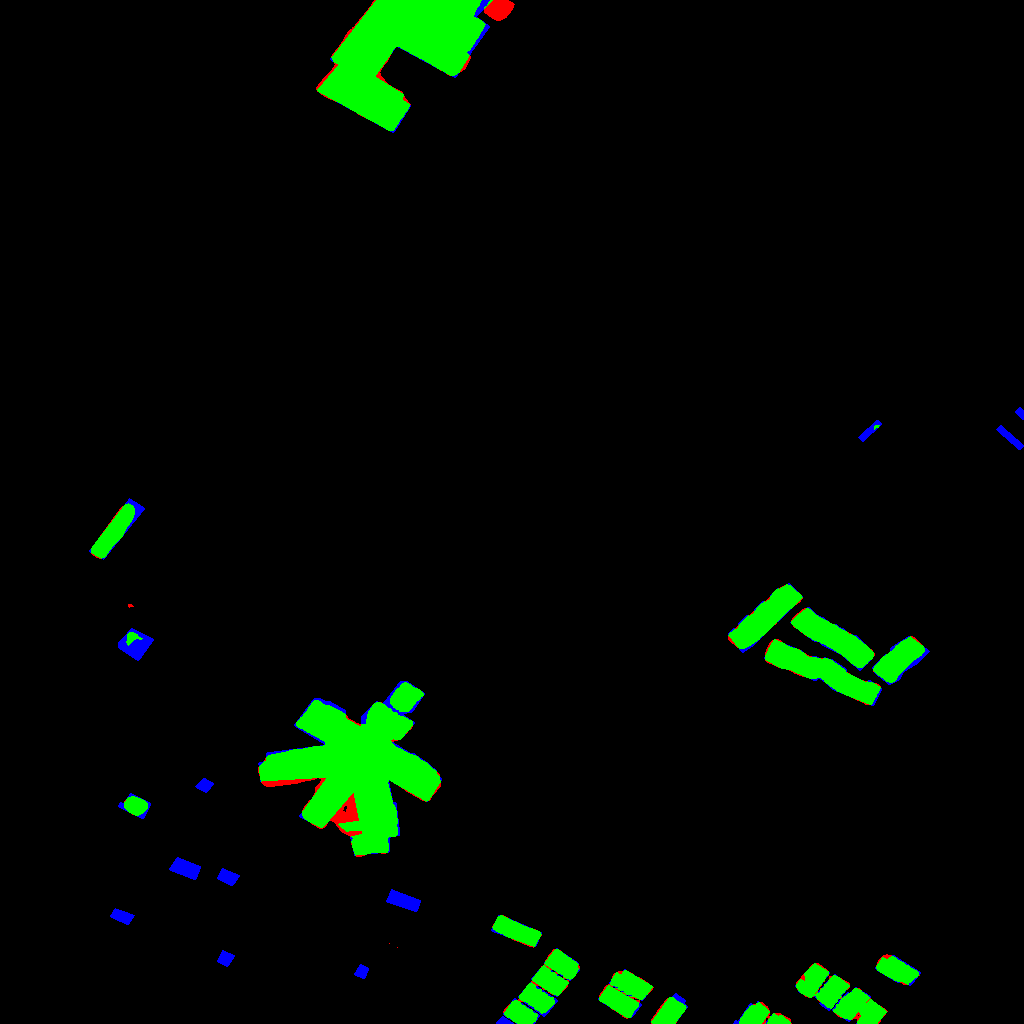}
      \end{minipage}
   }
   \subfigure[DPCC]{
      \begin{minipage}[b]{0.15\linewidth}
         \includegraphics[width=\linewidth]{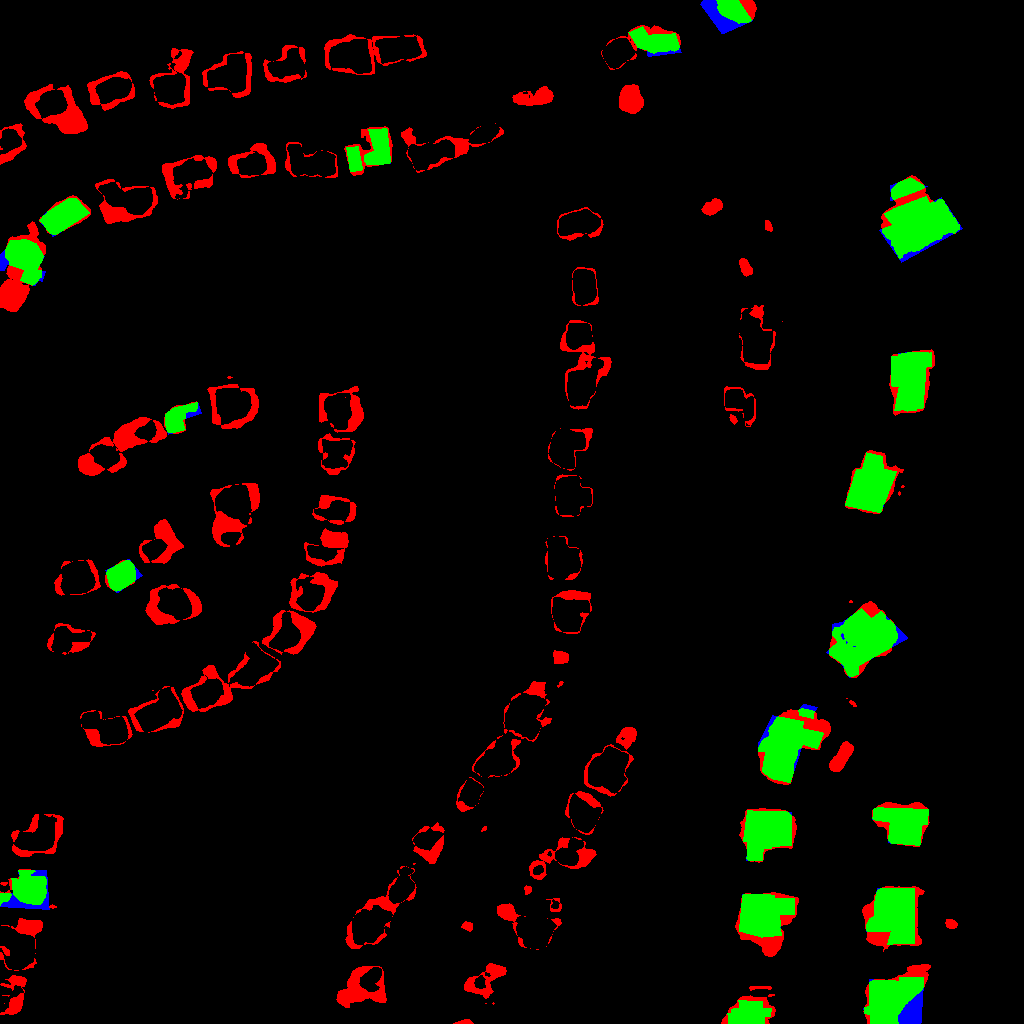}\vspace{4pt}
         \includegraphics[width=\linewidth]{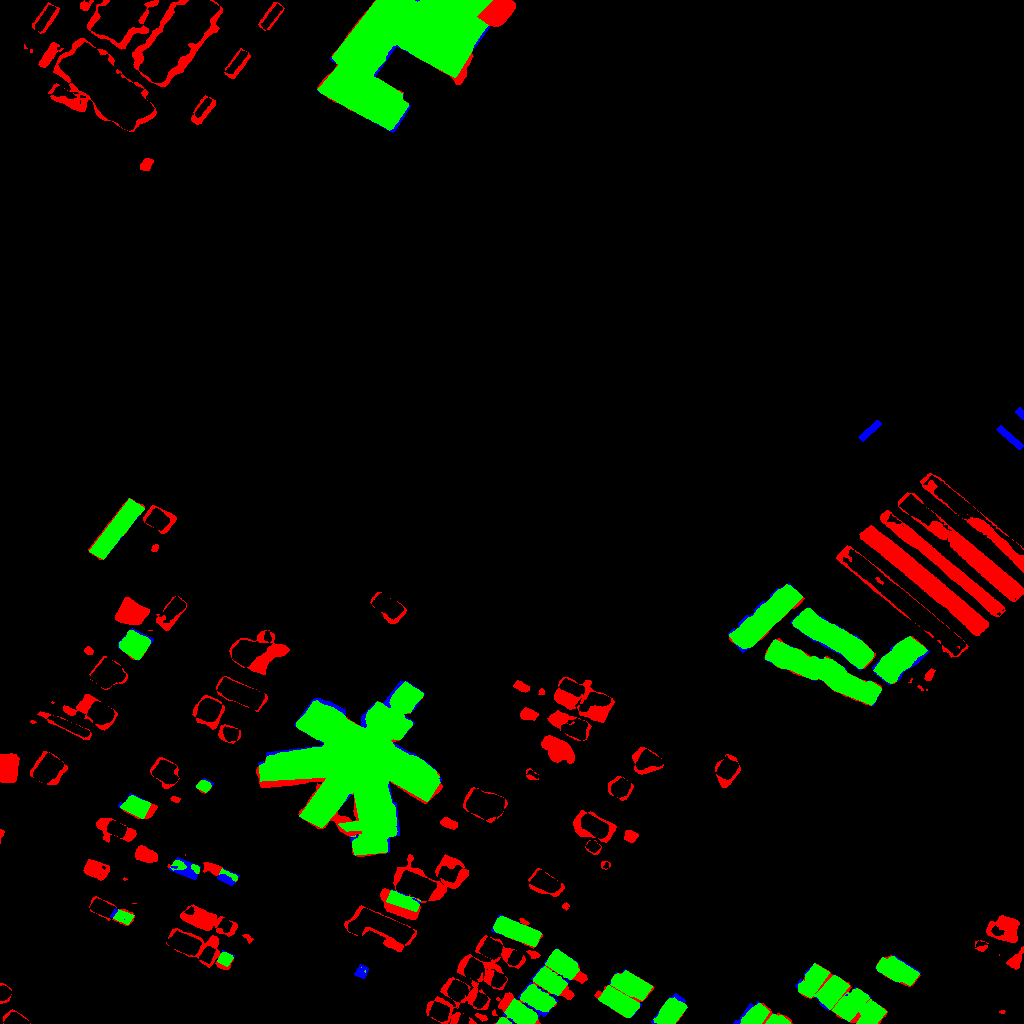}
      \end{minipage}
   }
   \subfigure[STAR]{
      \begin{minipage}[b]{0.15\linewidth}
         \includegraphics[width=\linewidth]{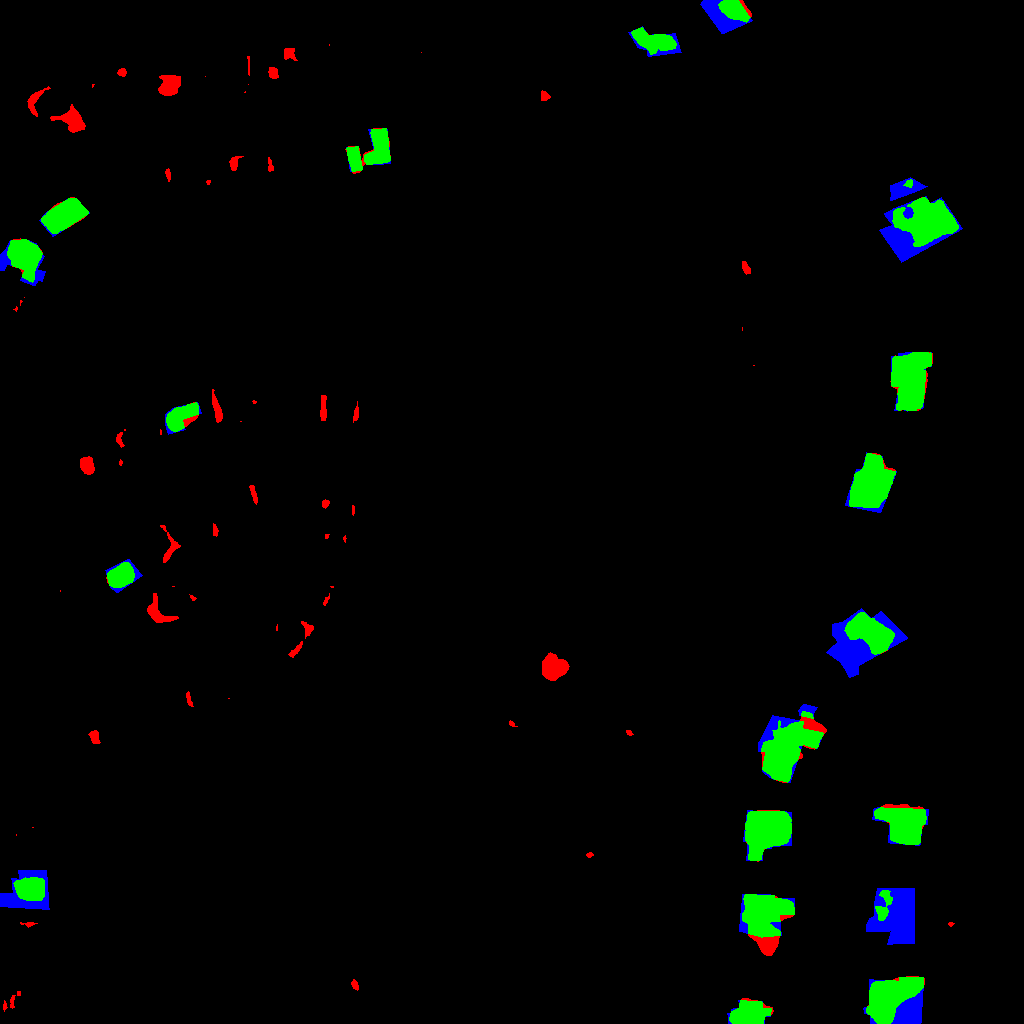}\vspace{4pt}
         \includegraphics[width=\linewidth]{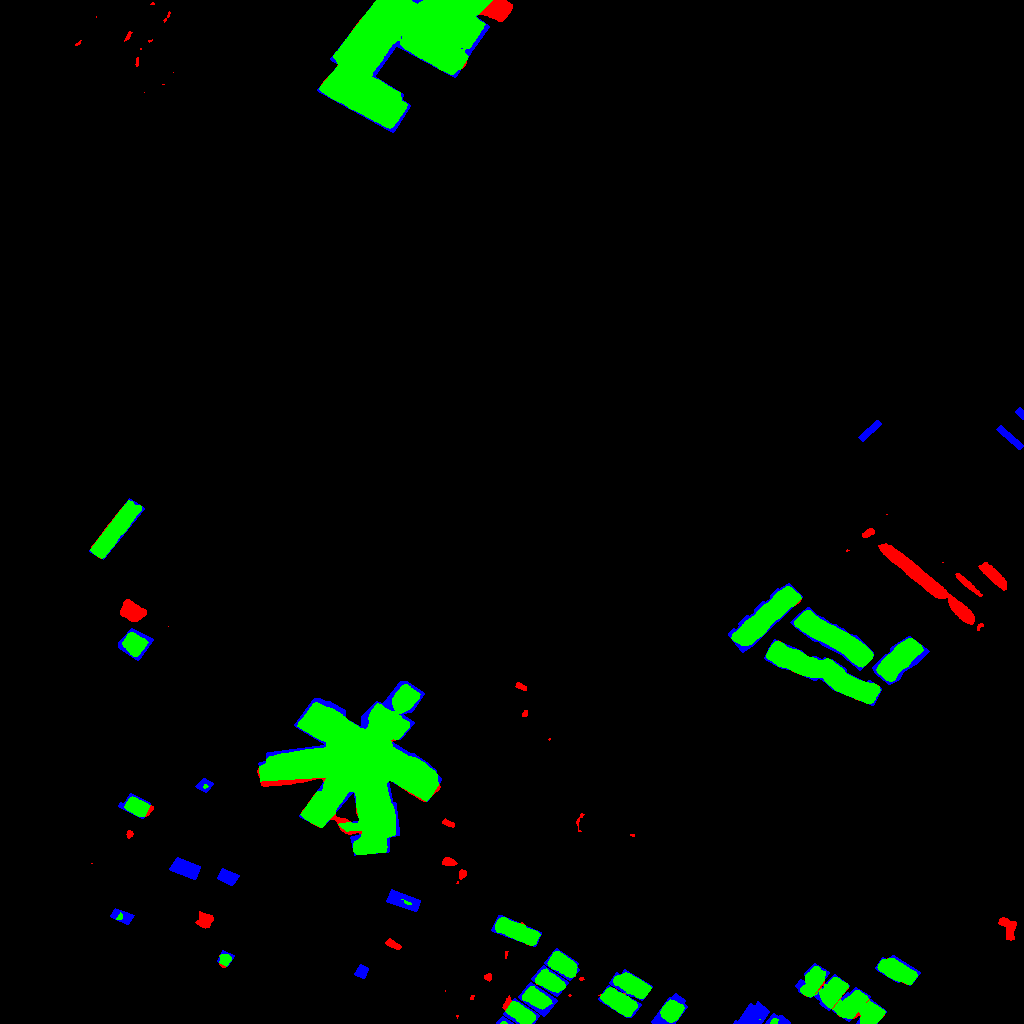}
      \end{minipage}
   }
   \caption{Error analysis for ChangeStar with bitemporal supervision, DPCC and ChangeStar with STAR.
      The basic segmentation model is FarSeg with ResNeXt-101 32x4d.
      The rendered colors represent \textcolor[rgb]{0,1,0}{true positives (TP)}, \textcolor[rgb]{1,0,0}{false positives (FP)}, and \textcolor[rgb]{0,0,1}{false negatives (FN)}.
   }
   \label{fig:error_maps}
\end{figure*}

\bfsec{Bitemporal Sup. vs. Single-Temporal Sup.}
To investigate the gap between bitemporal supervision and single-temporal supervision, we conducted comprehensive experiments to analyze their performance difference.
The results are presented in Table~\ref{tab:as_bi_st}.
We observe a 16$\sim$19\% F$_1$ gap between DPCC and bitemporal supervised methods when we used the 100\% xView2 pre-disaster dataset.
Our STAR can significantly bridge the gap to within 10\% when using a large backbone.
And it can be seen that the performance gap keeps getting smaller as the backbone network goes deeper.
To align the number of labeled images with the LEVIR-CD training set (445 pairs) for a more fair comparison, we subsampled the xView2 pre-disaster dataset into a subset of 890 images (~9.7\%).
We can find that when using fewer training data, ChangeStar still outperforms DPCC by a large margin.
As the number of training data decreases, the accuracy of ChangeStar meets degenerations in different degrees.
With a small backbone (R-18), the F$_1$ gap degenerates from -12.8\% to -20.6\%.
When we used a large backbone (RX-101 32x4d), this degeneration obtained improvement only from -9.7\% to -11.5\%.
This suggests that a large vision backbone can help single-temporal change representation learning in the cases of using sufficient and insufficient single-temporal training data.

\begin{table}[htb]
   \caption{Bitemporal supervision versus single-temporal supervision.
      All methods were evaluated on LEVIR-CD$^{\texttt{test}}$ for consistent comparison.
      \label{tab:as_bi_st}}
   \centering
   \renewcommand{\arraystretch}{1.5}
   \resizebox{\linewidth}{!}{
      \begin{tabular}{l|l|l|ccc}
         Method                              & Backbone     & Training data               & IoU (\%) & F$_1$ (\%) & F$_1$ gap (\%)$\downarrow$ \\ \shline
         \multicolumn{2}{l|}{\textit{Bitemporal Supervised}}
                         &                             &          &            &                \\
         ChangeStar                & R-18         & LEVIR-CD$^{\texttt{train}}$ & 82.3     & 90.2       & -              \\
         ChangeStar                  & R-50         & LEVIR-CD$^{\texttt{train}}$ & 83.1     & 90.8       & -              \\
         ChangeStar                  & RX-101 32x4d & LEVIR-CD$^{\texttt{train}}$ & 83.9     & 91.2       & -              \\
         \hline
         \multicolumn{2}{l|}{\textit{Single-Temporal Supervised}}             &                             &          &            &                \\
         DPCC                   & R-18         & xView2 pre.(100\%)         & 56.6     & 72.3       & -17.9          \\
         DPCC                   & R-50         & xView2 pre.(100\%)         & 55.8     & 71.7       & -19.1          \\
         DPCC                   & RX-101 32x4d & xView2 pre.(100\%)         & 59.5     & 74.6       & -16.6          \\
         \hline
         ChangeStar                  & R-18         & xView2 pre.(100\%)    & 63.2     & 77.4       & -12.8          \\
         ChangeStar                 & R-50          & xView2 pre.(100\%)    & 66.9     & 80.2       & -10.6          \\
         ChangeStar                  & RX-101 32x4d & xView2 pre.(100\%)    & 68.8     & 81.5       & -9.7           \\
         \hline
         DPCC                 & R-18         & xView2 pre.(890, 9.7\%)      & 45.5     & 62.5       & -27.7          \\
         DPCC                 & R-50         & xView2 pre.(890, 9.7\%)      & 48.2     & 65.0       & -25.8         \\
         DPCC                 & RX-101 32x4d & xView2 pre.(890, 9.7\%)      & 53.2     & 69.4       & -21.8          \\
         \hline
         ChangeStar                 & R-18         & xView2 pre.(890, 9.7\%)      & 53.4     & 69.6       &  -20.6     \\
         ChangeStar                 & R-50         & xView2 pre.(890, 9.7\%)      & 60.3     & 75.2       &  -15.6     \\
         ChangeStar                 & RX-101 32x4d & xView2 pre.(890, 9.7\%)      & 66.2     & 79.7       &  -11.5     \\
      \end{tabular}
   }
\end{table}

\bfsec{Error analysis.}
Comparing Fig.~\ref{fig:error_maps} (e) with Fig.~\ref{fig:error_maps} (d) and (f), we can find that the error of DPCC mainly lies in false positives due to various object appearance and object geometric offsets.
This is because DPCC only depends on semantic prediction to compare.
To alleviate this problem, that bitemporal supervision directly learns how to compare from pairwise labeled data, while STAR learns how to compare from unpaired labeled data.
From Fig.~\ref{fig:error_maps} (d)/(f), STAR is partly impacted by false positives due to the complete absence of the actual negative samples, e.g. the same object at different times.
Nevertheless, STAR can still learn helpful object change representation to recognize many unseen negative examples successfully.

\begin{figure}
   \begin{center}
      \includegraphics[width=0.9\linewidth]{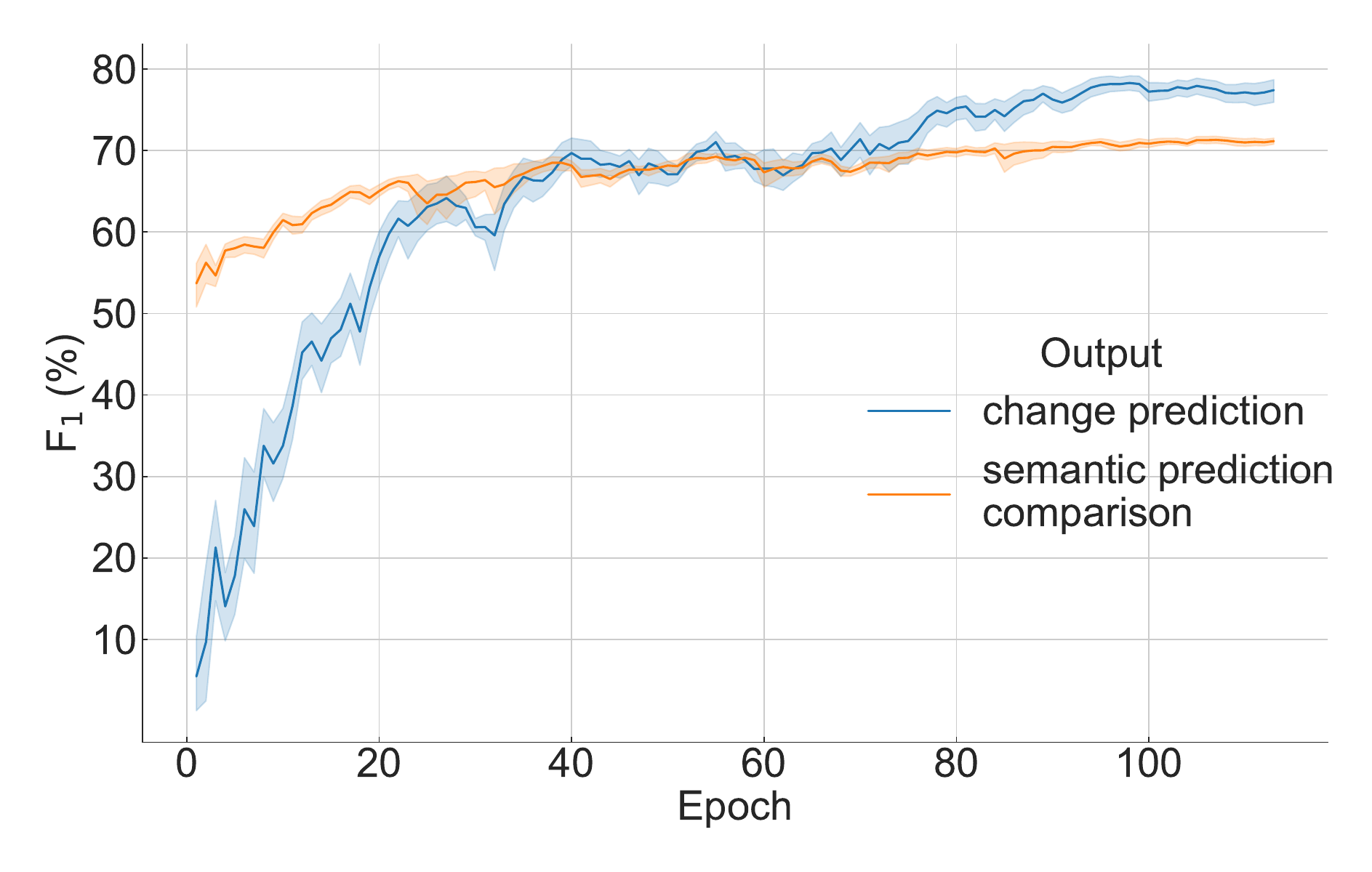}
   \end{center}
   \caption{Learning curves of IoU (\%) and F$_1$ (\%) on LEVIR-CD$^{\texttt{all}}$ using multi-task outputs from ChangeStar with FarSeg.
   The multi-task outputs include a change prediction from ChangeMixin and two semantic predictions from FarSeg.
   }
   \label{fig:multitask_output}
\end{figure}

\bfsec{Does STAR really work?}
ChangeStar can simultaneously output bitemporal semantic predictions and the change prediction.
The change prediction can also be obtained by semantic prediction comparison.
We thus show their learning curves to explore their relationship, as shown in Fig.~\ref{fig:multitask_output}.
We find that the semantic representation learning has a faster convergence speed than the object change representation learning in ChangeStar.
In the early stage of training ($(0, 40]$ epochs), semantic prediction comparison is superior to change prediction.
This suggests that learning semantic representation is easier than learning object change representation.
In the middle stage ($(40, 60]$ epochs), change prediction achieves similar performance with semantic prediction comparison.
After model convergence, change prediction performs better than semantic prediction comparison with a large margin.
This observation suggests that STAR can bring extra contrastive information to assist object change representation learning rather than only benefit from semantic supervision.

\bfsec{ChangeStar2 \textit{v.s.} ChangeStar.}
To illustrate step-by-step improvement from ChangeStar to ChangeStar2, we conduct an ablation study using ChangeStar with R-18 as the starting point, as presented in Table~\ref{tab:ablation_changestar_v2}.
ChangeStar2 involves two improvements, i.e., the new architecture and the new sampling algorithm.
In terms of architecture, the temporal difference network is integrated into ChangeMixin via residual connection, boosting the performance by 1.5\% F$_1$ and 2.0\% IoU.
In terms of sampling algorithm, stochastic self-contrast significantly improves the performance by 3.8\% F$_1$ and 5.2\% IoU, suggesting that synthesizing next-time image via single-temporal image is a very promising way to approximate real-world bitemporal image space.

\begin{table}[htb]
   \caption{Ablation study for each component extra introduced in ChangeStar2.
      The single-temporal supervised results are reported on LEVIR-CD$^{\texttt{test}}$ for consistent comparison.
      \label{tab:ablation_changestar_v2}}
   \centering
   \renewcommand{\arraystretch}{1.5}
   \resizebox{\linewidth}{!}{
      \begin{tabular}{l|l|ll}
         Method                               & Training data       & IoU (\%)      & F$_1$ (\%)   \\ \shline
         ChangeStar                           & xView2 pre-disaster & 63.2          & 77.4         \\
         + Temporal Difference Network        & xView2 pre-disaster & 65.2\up{2.0}  & 78.9\up{1.5} \\
         + Stochastic Self-Contrast           & xView2 pre-disaster & 68.4\up{5.2}  & 81.2\up{3.8} \\
      \end{tabular}
   }
\end{table}

\begin{figure}[htb]
   \centering
   \subfigure[Learning curves of ChangeMixin and ChangeMixin2]{
      \begin{minipage}[b]{\linewidth}
         \includegraphics[width=0.96\linewidth]{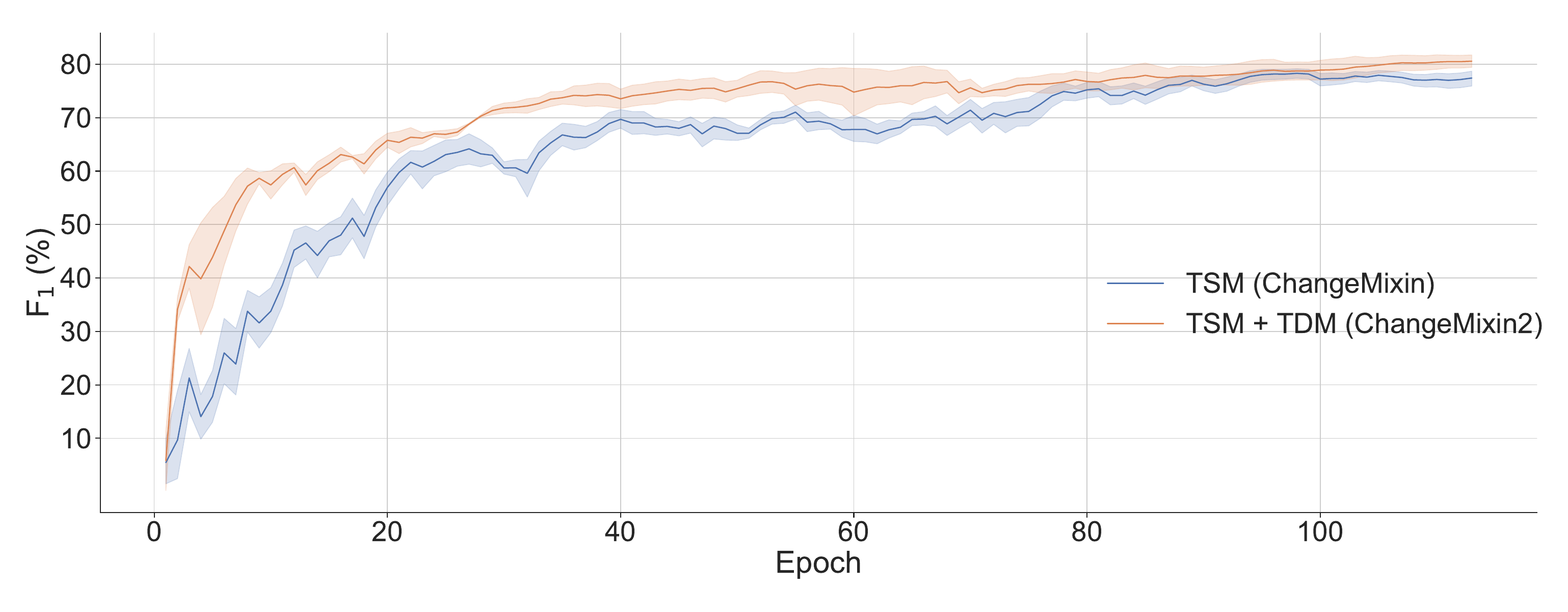}
      \end{minipage}
   }
   \subfigure[Positive and negative components of BCE loss]{
      \begin{minipage}[b]{\linewidth}
         \includegraphics[width=\linewidth]{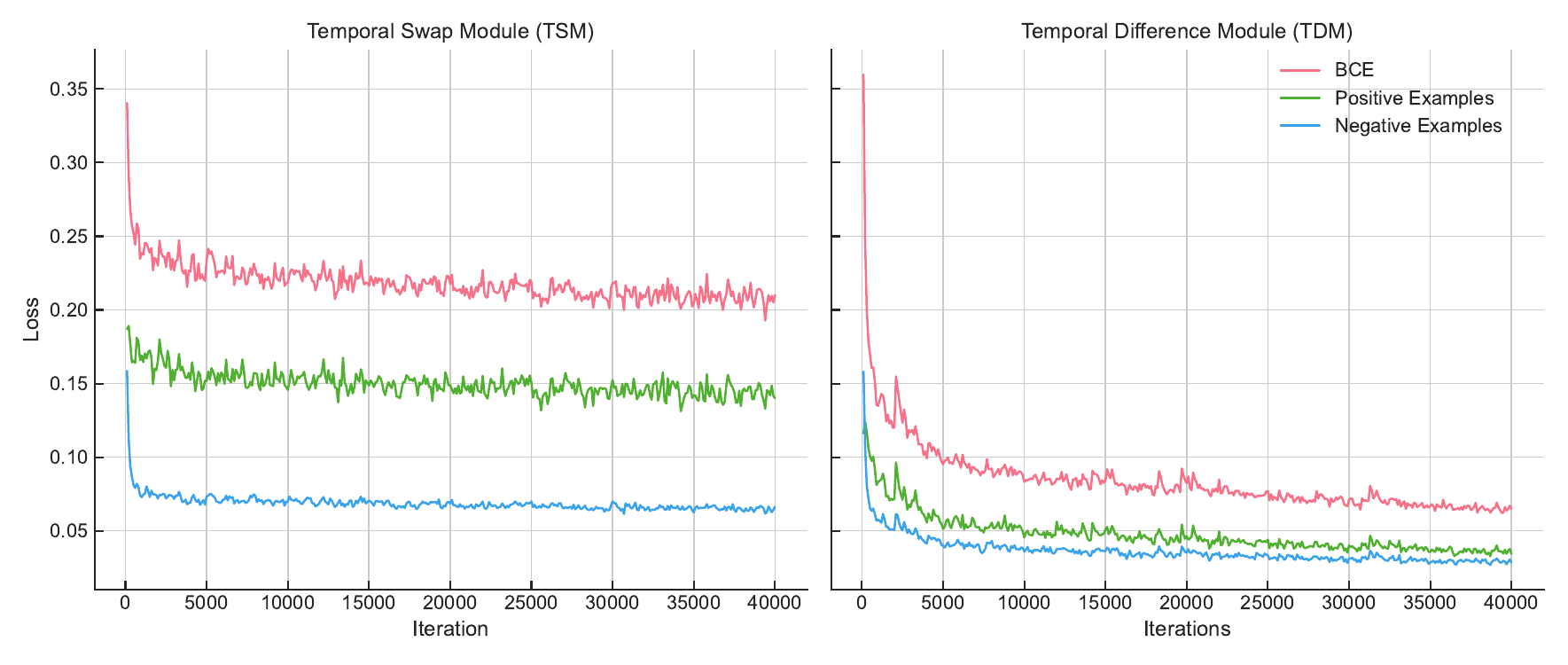}
      \end{minipage}
   }
   \subfigure[Average BCE loss for positive and negative examples]{
      \begin{minipage}[b]{\linewidth}
         \includegraphics[width=\linewidth]{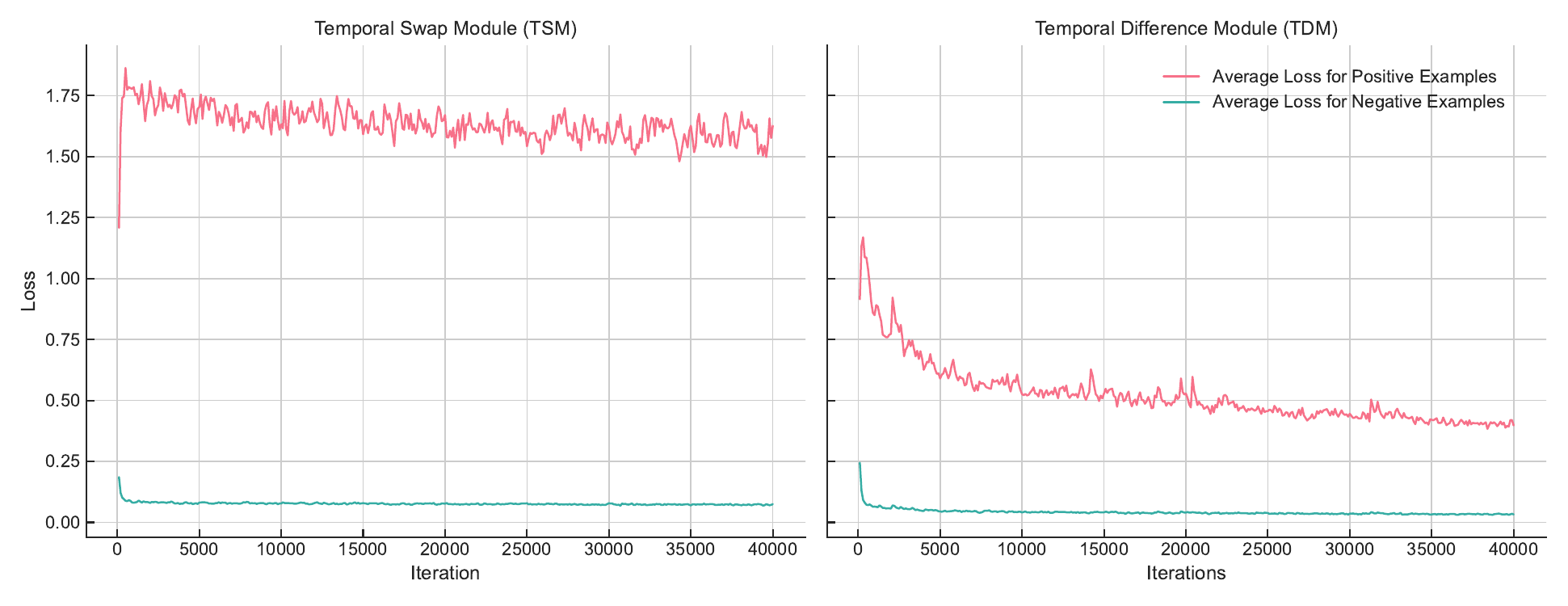}
      \end{minipage}
   }
   \caption{\textbf{Temporal difference module \textit{v.s.} Temporal swap module} via learning curves and loss curves.
   }
   \label{fig:tdm_tsm}
\end{figure}

\bfsec{How does TDM alleviate slow early convergence?}
As shown in Fig.~\ref{fig:tdm_tsm} (a), temporal difference module (TDM) significantly speeds up early convergence.
In addition to knowing TDM works well, we further investigate why it works from the perspective of optimization.
We first compare BCE loss curves of TSM and TDM in Fig.~\ref{fig:tdm_tsm} (b) and utilize the descent rate of loss value to measure the convergence rate approximately.  
It is obviously observed that TDM has a much faster convergence rate than TSM during the first 5k iterations since TDM provides a learning-free change representation.
Besides, we show BCE losses of positive and negative examples to observe their respective contributions.
This result suggests that the main obstacle to learning change representation in TSM is optimizing positive examples, while TDM, as a learning-free surrogate, can successfully overcome this problem.
We further measure average optimization difficulty for positive and negative examples via computing average loss, as shown in Fig.~\ref{fig:tdm_tsm} (c).
We find that the optimization of positive examples is obviously harder than negative examples for both TSM and TDM.
The initial smaller loss value of TDM confirms that a learning-free surrogate is an effective starting point for change representation learning.
Besides, it is observed that TDM has a larger descent magnitude and smaller variance of loss values for positive examples, suggesting that an effective starting point can stabilize subsequent change representation learning.

\begin{figure}[ht]
   \begin{center}
      \includegraphics[width=\linewidth]{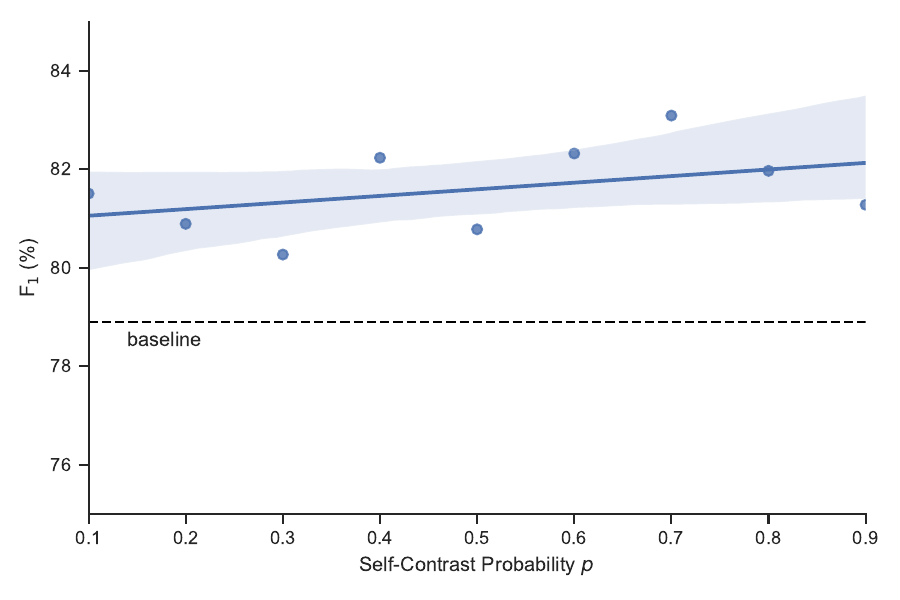}
   \end{center}
   \caption{Ablation study for self-contrast probability $p$ in stochastic self-contrast.
   }
   \label{fig:ab_sc_p}
\end{figure}

\begin{table}[!htb]
   \caption{Ablation study for stronger augmentation as an approximation in stochastic self-contrast in ChangeStar2.
      The single-temporal supervised results are reported on LEVIR-CD$^{\texttt{test}}$ for consistent comparison.}
      \label{tab:ablation_strong_aug}
   \centering
   \renewcommand{\arraystretch}{1.5}
   \resizebox{\linewidth}{!}{
      \begin{tabular}{l|l|ll}
         Method                                   & IoU (\%)      & F$_1$ (\%)   \\ \shline
         Baseline                                 & 65.2  & 78.9 \\
         + Stochastic Self-Contrast (default)     & 68.4  & 81.2 \\
         + Stochastic Self-Contrast (strong)      & 71.1  & 83.2 \\
      \end{tabular}
   }
\end{table}

\bfsec{Probability $p$ for Stochastic Self-Contrast.}
Self-contrast probability $p$ indicates the possibility of replacing the pseudo next-time image with an augmented copy.
Larger $p$ means a potentially higher ratio of such negative examples.
Fig.~\ref{fig:ab_sc_p} shows the relationship between self-contrast probability $p$ and F$_1$.
We can find the selection of $p$ is robust to the final F$_1$, which are all higher than the baseline.
Considering the positive correlation between self-contrast probability $p$ and F$_1$ is significant ($p{\rm -value}<0.05$), we choose self-contract $p$ of 0.9 as the default.

\bfsec{Exploring Stronger Augmentation as Approximation in Stochastic Self-Contrast.}
The authentic negative examples always include seasonal and radiation differences caused by diverse satellite sensor imaging conditions (e.g., atmosphere, weather) and object appearance changes as season changes.
We can find these distortions mainly belong to the color transformation; therefore, we use random color jitter as the default approximation.
Here, we explore stronger augmentation as an approximation, including random tone curve, sun flare, shadow, fog, rain, snow, cutout, Gaussian noise, and blur.
Table~\ref{tab:ablation_strong_aug} suggests that stronger augmentation as an approximation is promising, further improving by 2\% F$_1$.
This means that diverse negative examples are essential for single-temporal supervised learning. 
A more advanced approximation method is a promising direction to improve single-temporal supervised learning.

\begin{table}[htb]
   \caption{\textbf{Model Scaling} significantly reduces the performance gap between bitemporal supervision and single-temporal supervision.
      All methods were evaluated on LEVIR-CD$^{\texttt{test}}$ for consistent comparison.
      \label{tab:model_scaling}}
   \centering
   \renewcommand{\arraystretch}{1.5}
   \resizebox{\linewidth}{!}{
      \begin{tabular}{l|l|l|ccc}
         Method                              & Backbone     & Training data               & IoU (\%) & F$_1$ (\%) & F$_1$ gap (\%)$\downarrow$ \\ \shline
         \multicolumn{2}{l|}{\textit{Bitemporal Supervised}}
                         &                             &          &            &                \\
         ChangeStar2               & Swin-T       & LEVIR-CD$^{\texttt{train}}$ & 83.2     & 90.8       & -              \\
         ChangeStar2               & Swin-B       & LEVIR-CD$^{\texttt{train}}$ & 83.3     & 90.9       & -              \\
         ChangeStar2               & Swin-L       & LEVIR-CD$^{\texttt{train}}$ & 83.6     & 91.1       & -              \\
         \hline
         \multicolumn{2}{l|}{\textit{Single-Temporal Supervised}}             &                             &          &            &                \\
         DPCC                   & Swin-T       & xView2 pre-disaster         & 59.8     & 74.8       & -16.0          \\
         DPCC                   & Swin-B       & xView2 pre-disaster         & 62.3     & 76.7       & -14.2          \\
         DPCC                   & Swin-L       & xView2 pre-disaster         & 64.1     & 78.1       & -13.0          \\
         \hline
         ChangeStar2                & Swin-T       & xView2 pre-disaster         & 70.8     & 82.9       & -7.9           \\
         ChangeStar2                & Swin-B       & xView2 pre-disaster         & 72.5     & 84.1       & -6.8           \\
         ChangeStar2                & Swin-L       & xView2 pre-disaster         & 74.5     & 85.4       & -5.7           \\
      \end{tabular}
   }
\end{table}

\bfsec{Transformer-based Model Scaling for ChangeStar2.}
The effectiveness of ConvNet-based model scaling for ChangeStar has been confirmed in Table~\ref{tab:as_bi_st}.
We further demonstrate the effectiveness of Transformer-based model scaling for ChangeStar2.
Table~\ref{tab:model_scaling} presents that by scaling up the backbone from Swin-T to Swin-L, the performances of all entries obtain improvements.
In particular, single-temporal supervised ChangeStar2 improves F$_1$ from 82.9\% to 85.4\%, significantly reducing the performance gap with bitemporal supervised models to only -5.7\%.
We guess that the performances of bitemporal supervised ChangeStar2 models have achieved saturation on the LEVIR-CD dataset, approximately 91\% F$_1$. 
Because ChangeStar with RX-101 32x4d has achieved 91.2\% F$_1$, while ChangeStar2 with Swin-L, such a big model, only achieves 91.1\% in this approximate i.i.d scenario.
However, single-temporal supervised ChangeStar2 still improved substantially in the cross-domain scenario, which confirms that STAR is very promising to become a necessary label-efficient learning paradigm in real-world applications at scale.

\subsection{Discussion: Advantages and Limitations of STAR}
\subsubsection{Advantages and Potentials of STAR}
\bfsec{STAR simplifies the image collection.}
Image data collection is the first step to preparing supervision.
However, it is non-trivial to collect image pairs that include changes since the change is naturally sparse.
Not every image pair has a chance to be a sample for change detection.
STAR addresses this problem by learning change representation from unpaired images, thus simplifying the image collection stage in practical applications.

\begin{table}[ht]
\caption{Performance summary of single-temporal supervised learning (STAR) compared to bitemporal supervised learning (BSL).
The network architectures are the same for STAR and BSL.
\clegend{indomain}{in-domain} and \clegend{outdomain}{out-of-domain} represent that the scenario of this result, respectively.
\label{tab:summary}}
\centering
\renewcommand{\arraystretch}{1.5}
\resizebox{\linewidth}{!}{
\begin{tabular}{l|cccccc|cccc}
       & \multicolumn{4}{c|}{DynamicEarthNet}           & \multicolumn{2}{c|}{WHU-CD} & \multicolumn{2}{c|}{LEVIR-CD}    & \multicolumn{2}{c}{S2Looking} \\ \cline{2-11} 
Method & SCS  & BC   & SC   & \multicolumn{1}{c|}{mIoU} & IoU (\%)          & F$_1$ (\%)           & IoU (\%)  & \multicolumn{1}{c|}{F$_1$ (\%)}   & IoU (\%)           & F$_1$ (\%)            \\ \shline
BSL    & \cellcolor[HTML]{DAE8FC}19.4 & \cellcolor[HTML]{DAE8FC}12.9 & \cellcolor[HTML]{DAE8FC}26.0 & \multicolumn{1}{c|}{\cellcolor[HTML]{DAE8FC}41.8} & \cellcolor[HTML]{DAE8FC}74.4 & \cellcolor[HTML]{DAE8FC}85.3 & \cellcolor[HTML]{DAE8FC}83.9 & \multicolumn{1}{c|}{\cellcolor[HTML]{DAE8FC}91.2} & \cellcolor[HTML]{DAE8FC}51.3 & \cellcolor[HTML]{DAE8FC}67.8 \\
STAR   & \cellcolor[HTML]{DAE8FC}20.2 & \cellcolor[HTML]{DAE8FC}15.1 & \cellcolor[HTML]{DAE8FC}25.3 & \multicolumn{1}{c|}{\cellcolor[HTML]{DAE8FC}42.1} & \cellcolor[HTML]{DAE8FC}72.8 & \cellcolor[HTML]{DAE8FC}84.3 & \cellcolor[HTML]{FFE38E}66.2 & \multicolumn{1}{c|}{\cellcolor[HTML]{FFE38E}79.7} & \cellcolor[HTML]{FFE38E}19.5 & \cellcolor[HTML]{FFE38E}32.7 
\end{tabular}}
\end{table}

\bfsec{STAR reduces the difficulty of data annotation.}
Change region annotation belongs to a multi-temporal image data annotation, whose labeling pipeline is more complex, generally including seeking change regions, drawing polygon regions, and double-checking.
Its difficulty and time bottleneck mainly lie in seeking change regions and double-checking rather than drawing, which is the intensive operation of single image annotation.
This is because single image annotation is always ``what you see is what you get''.
STAR only needs single image annotation, thus skipping two difficulties of change data annotation, i.e., seeking change regions and double-checking.

\bfsec{Open map produces can reduce labeling cost for STAR.}
In geoscience and remote sensing communities, many products can significantly reduce single-temporal image labeling costs. 
There are many public and private single-temporal land-use/land-cover products, such as Open Street Map and national survey data of land use and natural resources. 
These data can accelerate the labeling process for single-temporal tasks, such as semantic segmentation object detection. 
However, extra cloud-free image collection, careful data cleaning, and human checking must be used to produce labels for multi-temporal tasks like change detection.
If we use single-temporal supervision, these manual costs can be avoided in the whole training data production pipeline.

\subsubsection{Limitations and Suitable scenarios of STAR}
Compared to bitemporal supervised learning (BSL), STAR still has some limitations.
To investigate these limitations and suitable application scenarios of STAR, We summarize the performance of STAR and BSL in Table~\ref{tab:summary}.
The network architectures are the same for STAR and BSL.
We observe that STAR works well in in-domain scenarios, achieving better results on the DynamicEarthNet dataset or competitive results on the WHU-CD dataset compared to BSL.
This is because pseudo-bitemporal image pairs can bring more supervisory signals to promote learning the semantic dissimilarity of objects.
These results also reflect that out-of-domain generalization seems to be an open problem for STAR.
In out-of-domain cases, STAR's training data adopt the xView2 pre-diaster dataset, and the test data adopt the LEVIR-CD and S2Looking datasets.
Apart from the difference in the learning algorithm itself, there is a large domain gap between single-temporal training data and bitemporal test data, especially when different temporal data has different domains (e.g., pre/post-disaster, large variation in off-nadir).
This domain gap causes STAR's performance not to be as good as it can achieve in in-domain scenarios.
How to address this out-of-domain generalization problem and how to build a more suitable benchmark (including dataset, scenario design, and label cost analysis) for STAR are great valuable future works.

\section{Conclusion}
\label{sec:conc}

We present single-temporal supervised learning (STAR) for universal remote sensing change detection to bypass the problem of collecting labeled image pairs in conventional bitemporal supervised learning.
STAR provides a new perspective of exploiting changes of arbitrary image pairs in the entire bitemporal image space as the supervisory signals to learn generalizable change representation.
To demonstrate the flexibility and scalability of STAR, we design a simple yet unified change detection architecture, termed ChangeStar2, capable of addressing binary change detection, object change detection, and semantic change detection.
The extensive experimental analysis shows its superior performance in different domains.
We hope that STAR will serve as a solid and strong baseline and help ease future research in single-temporal supervised change detection.

\smallsec{Data Availability Statements} The datasets analyzed during this study are all publicly available for research purpose - the \href{https://xview2.org/}{xView2}, \href{https://spacenet.ai/sn8-challenge/}{SpaceNet 8}, \href{https://justchenhao.github.io/LEVIR/}{LEVIR-CD}, \href{https://gpcv.whu.edu.cn/data/building_dataset.html}{WHU-CD}, \href{https://mediatum.ub.tum.de/1650201}{DynamicEarthNet}, \href{https://drive.google.com/file/d/1GX656JqqOyBi_Ef0w65kDGVto-nHrNs9}{CDD}, \href{https://github.com/S2Looking/Dataset}{S2Looking}, and \href{https://drive.google.com/file/d/1QlAdzrHpfBIOZ6SK78yHF2i1u6tikmBc/view?usp=sharing}{SECOND} datasets.



%
%



\bibliographystyle{spbasic}      
\bibliography{references}   

\end{document}